\documentclass[10pt,journal,compsoc]{IEEEtran}
\ifCLASSOPTIONcompsoc
  \usepackage[nocompress]{cite}
\else
  \usepackage{cite}
\fi
\ifCLASSINFOpdf
\else
\fi
\usepackage{graphicx}
\usepackage{subfigure}
\usepackage{amsmath}
\usepackage{amssymb}
\usepackage{tabularx}
\usepackage{verbatim}
\usepackage{url} 
\usepackage{pbox}

\usepackage[caption=false,font=footnotesize,labelfont=sf,textfont=sf]{subfig}
\begin{document}
%
\title{DeepIGeoS: A Deep Interactive Geodesic Framework for Medical Image Segmentation}
\author{Guotai Wang, Maria A. Zuluaga, Wenqi Li, Rosalind Pratt, Premal A. Patel, Michael Aertsen, Tom Doel, Anna L. David, Jan Deprest, S\'ebastien Ourselin, Tom Vercauteren
	\IEEEcompsocitemizethanks{\IEEEcompsocthanksitem G. Wang, M.A. Zuluaga, W. Li, R. Pratt, P.A. Patel, T. Doel, S. Ourselin and T. Vercauteren are with Translational Imaging Group, Wellcome EPSRC Centre for Interventional and Surgical Sciences (WEISS), University College London. R. Pratt and A.L. David are with Institute for Women's Health, University College London. M. Aertsen is with Department of Radiology, University Hospitals KU Leuven. J. Deprest is with Department of Obstetrics, University Hospitals KU Leuven.\protect\\
		E-mail: guotai.wang.14@ucl.ac.uk}
	}
\IEEEtitleabstractindextext{%
\begin{abstract}
Accurate medical image segmentation is essential for diagnosis, surgical planning and many other applications. Convolutional Neural Networks (CNNs) have become the state-of-the-art automatic segmentation methods. However, fully automatic results may still need to be refined to become accurate and robust enough for clinical use. We propose a deep learning-based interactive segmentation method to improve the results obtained by an automatic CNN and to reduce user interactions during refinement for higher accuracy. We use one CNN to obtain an initial automatic segmentation, on which user interactions are added to indicate mis-segmentations. Another CNN takes as input the user interactions with the initial segmentation and gives a refined result. We propose to combine user interactions with CNNs through geodesic distance transforms, and propose a resolution-preserving network that gives a better dense prediction. In addition, we integrate user interactions as hard constraints into a back-propagatable Conditional Random Field. We validated the proposed framework in the context of 2D placenta segmentation from fetal MRI and 3D brain tumor segmentation from FLAIR images. Experimental results show our method achieves a large improvement from automatic CNNs, and obtains comparable and even higher accuracy with fewer user interventions and less time compared with traditional interactive methods.
\end{abstract}

}

\maketitle

\IEEEdisplaynontitleabstractindextext

%
\IEEEpeerreviewmaketitle

\IEEEraisesectionheading{\section{Introduction}\label{sec:introduction}}

%
%
%
%
\IEEEPARstart{S}{egmentation} of anatomical structures is an essential task for a range of medical image processing applications such as image-based diagnosis, anatomical structure modeling, surgical planning and guidance. 
Although automatic segmentation methods~\cite{Sharma2010} have been investigated for many years, they can rarely achieve sufficiently accurate and robust results to be useful for many medical imaging applications. This is mainly due to poor image quality (with noise, artifacts and low contrast), large variations among patients, inhomogeneous appearances brought by pathology, and variability of protocols among clinicians leading to different definitions of a given structure boundary. Interactive segmentation methods, which take advantage of users' knowledge of anatomy and the clinical question to overcome the challenges faced by automatic methods, are widely used for higher accuracy and robustness~\cite{Zhao2013}. 

Although leveraging user interactions helps to obtain more precise segmentation~\cite{Boykov2001, Xu1998,Grady2006a, Criminisi2008}, the resulting requirement for many user interactions increases the burden on the user. A good interactive segmentation method should require as few user interactions as possible, leading to interaction efficiency. Machine learning methods are commonly used to reduce user interactions. For example, GrabCut~\cite{Rother2004} uses Gaussian Mixture Models to represent color distributions. It requires the user to provide a bounding box around the object of interest for segmentation and allows additional scribbles for refinement. SlicSeg~\cite{Wang2016} employs Online Random Forests to segment a Magnetic Resonance Imaging (MRI) volume by learning from user-provided scribbles in only one slice. Active
learning is used in~\cite{Wang2014a} to actively select candidate regions for querying the user.

Recently, deep learning techniques with convolutional neural networks (CNNs) have achieved increasing success in image segmentation~\cite{Girshick2014, Long2014, Chen2015iclr}. 
They can find the most suitable features through automatic learning instead of manual design. 
By learning from large amounts of training data, CNNs have achieved state-of-the-art performance for automatic segmentation~\cite{Chen2015iclr,Havaei2016, Kamnitsas2017}. One of the most widely used CNNs is the Fully Convolutional Network (FCN)~\cite{Long2014}. It outputs the segmentation directly by computing forward propagation only once at the testing time.

Recent advances of CNNs for image segmentation mainly focus on two aspects. The first is to overcome the problem of reduction of resolution caused by repeated combination of max-pooling and downsampling. 
 Though some upsampling layers can be used to recover the resolution, this easily leads to blob-like segmentation results and low accuracy for tiny structures\cite{Long2014}. In \cite{Onvolutions2016, Chen2015iclr}, dilated convolution is proposed to replace some downsampling layers and it allows exponential expansion of the receptive field without the loss of resolution. However, the CNNs in \cite{Onvolutions2016,Chen2015iclr} keep three layers of pooling and downsampling therefore their output resolution is still reduced eight times compared with the input. The second aspect is to enforce inter-pixel dependency to get a spatially regularized result. 
 This helps to recover edge details and reduce noise in pixel classification. DeepLab~\cite{Chen2016deeplab} and DeepMedic~\cite{Kamnitsas2017} used fully connected Conditional Random Fields (CRFs) as a post-processing step. However, the parameters of these CRFs rely on manual tuning which is time consuming and may not ensure optimal values. It was shown in~\cite{Zheng2015a} that the CRF can be formulated as a Recurrent Neural Network (RNN) so that it can be trained end-to-end utilizing the back-propagation algorithm. However, this CRF constrains the pairwise potentials as Gaussian functions, which may be too restrictive for some complex cases, and this method does not apply automatic learning to all its parameters. Thus, using more freeform learnable pairwise potential functions and allowing automatic learning of all the parameters can potentially achieve better results.  

This paper aims to integrate user interactions into CNN frameworks to obtain accurate and robust segmentation of 2D and 3D medical images, and at the same time, we aim to make the interactive framework more efficient with a minimal number of user interactions by using CNNs. With the good performance of CNNs shown for automatic image segmentation tasks \cite{Girshick2014, Long2014, Chen2016deeplab, Havaei2016, Kamnitsas2017}, we hypothesize that they can reduce the number of user interactions for interactive image segmentation. However, only a few works have been reported to apply CNNs to interactive segmentation tasks \cite{Abdulkadir2016,Lin2016,Rajchl2016,Xu2016}. 

The contributions of this work are four-fold. 1). We propose a deep CNN-based interactive framework for 2D and 3D medical image segmentation. 
We use one CNN to get an initial automatic segmentation, which is refined by another CNN that takes as input the initial segmentation and user interactions; 
2). We present a new way to combine user interactions with CNNs based on geodesic distance maps that are used as extra channels of the input for CNNs. 
We show that using geodesic distance can lead to improved segmentation accuracy compared with using Euclidean distance; 3). We propose a resolution-preserving CNN structure which leads to a more detailed segmentation result compared with traditional CNNs with resolution loss, and 4). We extend the current RNN-based CRFs~\cite{Zheng2015a} for segmentation so that the back-propagatable CRFs can use user interactions as hard constraints and all the parameters of potential functions can be trained in an end-to-end way. We apply the proposed method to 2D placenta segmentation from fetal MRI and 3D brain tumor segmentation from fluid attenuation inversion recovery (FLAIR) images.

\section{Related Works}
\subsection{Image Segmentation based on CNNs}
Typical CNNs such as AlexNet~\cite{Krizhevsky2012}, GoogleNet~\cite{Szegedy2015}, VGG~\cite{Simonyan2015} and ResNet~\cite{He2015res} were originally designed for image classification tasks. Some early works adapted such networks for pixel labeling with patch or region-based methods \cite{Havaei2016, Girshick2014}. 
Such methods achieved higher accuracy than traditional methods that relied on hand-crafted features. However, they suffered from inefficiency for testing. FCNs~\cite{Long2014} take an entire image as input and give a dense segmentation. In order to overcome the problem of loss of spatial resolution due to multi-stage max-pooling and downsampling, it uses a stack of deconvolution (a.k.a. upsampling) layers and activation functions to upsample the feature maps. 
Inspired by the convolution and deconvolution framework of FCNs, a U-shape network (U-Net)~\cite{Hefny2015a} and its 3D version~\cite{Abdulkadir2016} were proposed 
for biomedical image segmentation. A similar network (V-Net)~\cite{Milletari2016} was proposed  to segment the prostate from 3D MRI volumes.

To overcome the drawbacks of successive max-pooling and downsampling that lead to a loss of feature map resolution, dilated convolution~\cite{Chen2015iclr, Onvolutions2016} was proposed to preserve the resolution of feature maps and enlarge the receptive field to incorporate larger contextual information. In~\cite{Ondruska2016}, a stack of dilated convolutions was used for object tracking and semantic segmentation. Dilated convolution has also been used for instance-sensitive segmentation~\cite{Dai2016} and action detection from video frames~\cite{Lea2016}.

Multi-scale features extracted from CNNs have been shown to be effective for improving segmentation accuracy~\cite{Long2014, Chen2015iclr,Onvolutions2016}. One way of obtaining multi-scale features is to pass several scaled versions of the input image through the same network. The features from all the scales can be fused for pixel classification~\cite{Lin2016cvpr_efficient}. 
 In~\cite{Havaei2016, Kamnitsas2017}, the features of each pixel were extracted from two concentric patches with different sizes. In~\cite{Pinheiro2014}, multi-scale images at different stages were fed into a recurrent convolutional neural network. Another widely used way to obtain multi-scale features is exploiting the feature maps from different levels of a CNN. For example, in~\cite{Hariharan2015}, features from intermediate layers are concatenated for segmentation and localization. In~\cite{Long2014, Chen2015iclr}, predictions from the final layer are combined with those from previous layers. 

\subsection{Interactive Image Segmentation}
Interactive image segmentation has been widely used in various applications~\cite{Armstrong2007,Cates2004,Rajchl2016}. There are many kinds of user interactions, such as click-based~\cite{Haider2015}, contour-based~\cite{Xu1998} and bounding box-based methods~\cite{Rother2004}. Drawing scribbles is user-friendly and particularly popular, e.g., in Graph Cuts~\cite{Boykov2001}, GeoS~\cite{Bai2007, Criminisi2008}, and Random Walks~\cite{Grady2006a}. 
However, most of these methods rely on low-level features and require a relatively large amount of user interactions to deal with images with low contrast and ambiguous boundaries. Machine learning methods~\cite{Barinova2012, Wang2016,Luengo2017} have been proposed to learn from user interactions. They can achieve higher segmentation accuracy with fewer user interactions. However, they are limited by hand-crafted features that depend on the user's experience.

Recently, using deep CNNs to improve interactive segmentation has attracted increasing attention due to CNNs' automatic feature learning  and high performance.  For instance, 3D U-Net~\cite{Abdulkadir2016} learns from sparsely annotated images and can be used for semi-automatic segmentation. ScribbleSup~\cite{Lin2016} also trains CNNs for semantic segmentation supervised by scribbles. DeepCut~\cite{Rajchl2016} employs user-provided bounding boxes as annotations to train CNNs for the segmentation of fetal MRI. However, these methods are not fully interactive for testing since they do not accept further interactions for refinement. In~\cite{Xu2016}, a deep interactive object selection method was proposed where user-provided clicks are transformed into Euclidean distance maps and then concatenated with the input of FCNs. However, the Euclidean distance does not take advantage of image context information. In contrast, the geodesic distance transform~\cite{Bai2007, Criminisi2008, Kohli2013} encodes spatial regularization and contrast-sensitivity but it has not been used for CNNs.  

\subsection{CRFs for Spatial Regularization}
Graphical models such as CRFs~\cite{Boykov2004,Yuan2010,Chen2015iclr} have been widely used to enhance segmentation accuracy by introducing spatial consistency. In~\cite{Boykov2004}, spatial regularization was obtained by minimizing the Potts energy with a min-cut/max-flow algorithm. In~\cite{Yuan2010}, the discrete max-flow problem was mapped to its continuous optimization formulation. Such methods encourage segmentation consistency between adjacent pixel pairs with high similarity.  In order to better model long-range connections within the image, a fully connected CRF was used in~\cite{Payet2010} to establish pairwise potentials on all pairs of pixels in the image. To make the inference of this CRF efficient, the pairwise edge potentials were defined by a linear combination of Gaussian kernels in~\cite{Krahenbuhl2011}. The parameters of CRFs in these works were manually tuned or inefficiently learned by grid search.  
In~\cite{Szummer2008}, a maximum margin learning method was proposed to learn CRFs using Graph Cuts. Other methods including structured output Support Vector Machines~\cite{Blaschko2014}, approximate marginal inference~\cite{Domke2013} and gradient-based optimization~\cite{Krahenbuhl2013a} were also proposed to learn parameters in CRFs. They treat the learning of CRFs as an independent step after the training of classifiers.

The CRF-RNN network~\cite{Zheng2015a} formulated dense CRFs as RNNs so that the CNNs and CRFs can be jointly trained in an end-to-end system for segmentation. However, the pair-wise potentials in~\cite{Zheng2015a} are limited to weighted Gaussians and not all the parameters are trainable due to the Permutohedral lattice implementation~\cite{Adams2010}. In~\cite{Vemulapalli2016}, a Gaussian Mean Field (GMF) network was proposed and combined with CNNs where all the parameters are trainable. More freeform pairwise potentials for a pair of super-pixels or image patches were proposed in~\cite{Liu2014,Lin2016cvpr_efficient}, but such CRFs have a low resolution. In~\cite{Kirillov2016}, a generic CNN-CRF model was proposed to handle arbitrary potentials for labeling body parts in depth images. However, it has not yet been validated with other segmentation applications.

\section{Method}

\begin{figure}[t]
	\centering
	\includegraphics[width=1.0\linewidth]{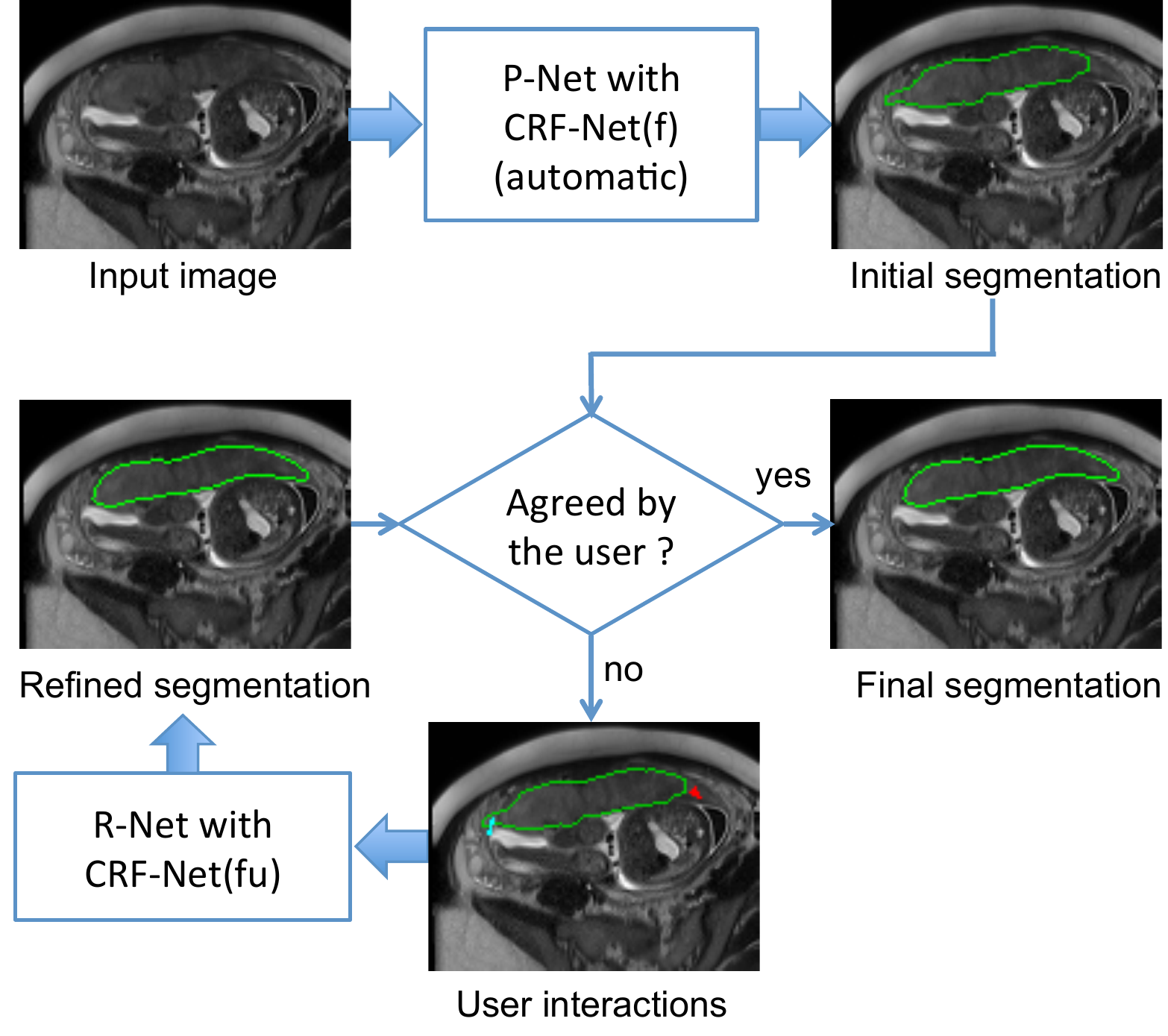}
	\caption[Overview of the proposed interactive segmentation method]{ 
		Overview of the proposed interactive segmentation method. P-Net proposes an initial automatic segmentation that is refined by R-Net with user interactions indicating mis-segmentations. 
		CRF-Net(f) is our proposed back-propagatable CRF that uses freeform pairwise potentials. It is extended to be CRF-Net(fu) that employs user interactions as hard constraints. } 
	\label{fig:framework}
\end{figure}

The proposed deep interactive segmentation method based on CNNs and geodesic distance transforms (DeepIGeoS) is depicted in Fig.~\ref{fig:framework}. To minimize the number of user interactions, we propose to use two CNNs: an initial segmentation proposal network (P-Net) and a refinement network (R-Net). P-Net takes as input a raw image with $C_I$ channels and gives an initial automatic segmentation. Then the user checks the segmentation and provides some interactions (clicks or scribbles) to indicate mis-segmented regions. R-Net takes as input the original image, the initial segmentation and the user interactions to provide a refined segmentation. 
P-Net and R-Net use a resolution-preserving structure that captures high-level features from a large receptive field without loss of resolution. They share the same structure except the difference in the input dimensions. Based on the initial automatic segmentation obtained by P-Net, the user might give clicks/scribbles to refine the result more than one time through R-Net. Differently from previous works~\cite{Wang2016a} that re-train the learning model each time when new user interactions are given, the proposed R-Net is only trained with user interactions once since it takes a considerable time to re-train a CNN model with a large training set.

\begin{figure}[t]
	\centering
	\includegraphics[width=1.0\linewidth]{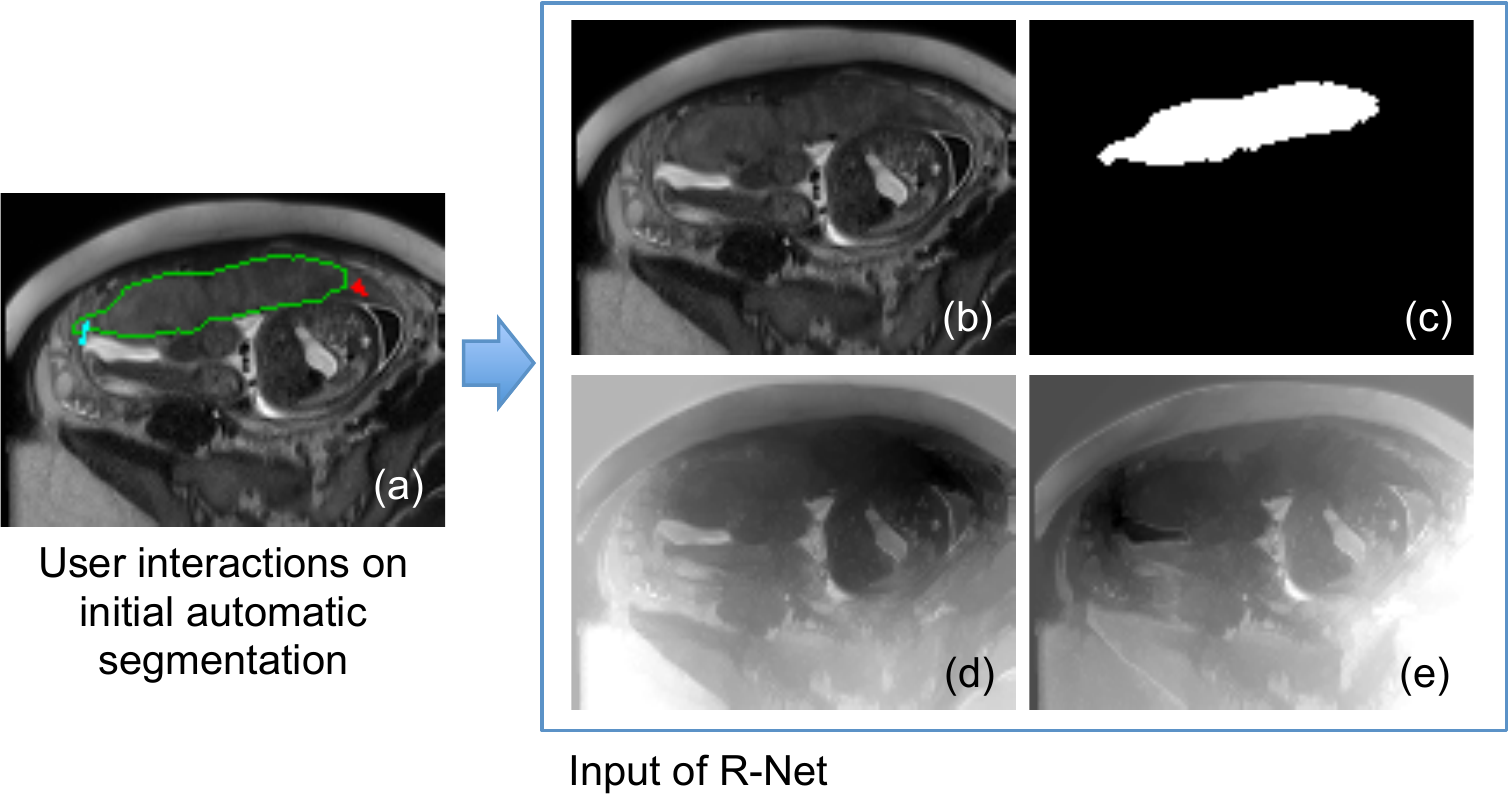}
	\caption[Input of R-Net using Geodesic distance transforms of user interactions]{ 
		Input of R-Net using geodesic distance transforms of user interactions. (a) The user provides clicks/scribbles to correct foreground(red) and background(cyan) on the initial automatic segmentation. (d) and (e) are geodesic distance maps based on foreground and background interactions, respectively. The original image (b) is combined with the initial automatic segmentation (c) and the geodesic distance maps (d), (e) by channel-concatenation and used as the input of R-Net.    } 
	\label{fig:scribble_distance}
\end{figure}
\begin{figure*}[t]
	\centering
	\includegraphics[width=1.0\linewidth]{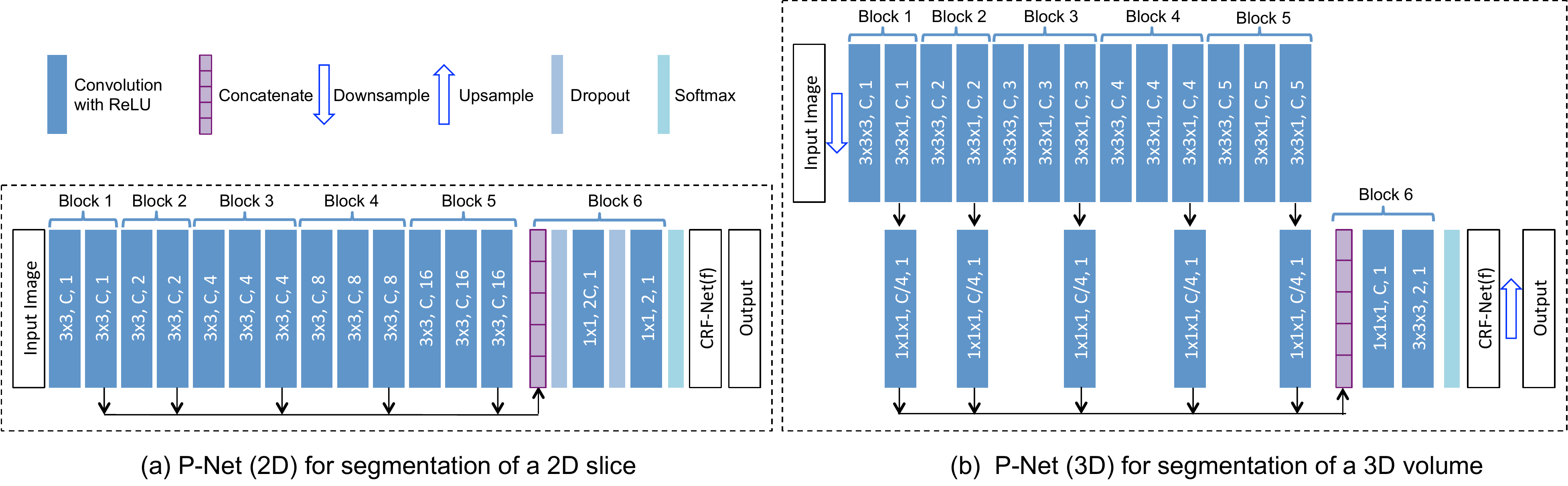}
	\caption[The CNN structure of 2D/3D P-Net with CRF-Net.]{ 
		The CNN structure of 2D/3D P-Net with CRF-Net(f). The parameters of convolution layers are (kernel size, output channels, dilation) in dark blue rectangles. Block 1 to block 6 are resolution-preserving. 2D/3D R-Net uses the same structure as 2D/3D P-Net except its input has three additional channels shown in Fig. \ref{fig:scribble_distance} and the CRF-Net(f) is replaced by the CRF-Net(fu) (Section~\ref{method:crf}).} 
	\label{fig:p-net}
\end{figure*}
To make the segmentation result more spatially consistent and to use scribbles as hard constraints, both P-Net and R-Net are connected with a CRF, which is modeled as an RNN (CRF-Net) so that it can be trained jointly with P-Net/R-Net by back-propagation. We use freeform pairwise potentials in the CRF-Net. The way user interactions are used is presented in~\ref{method:user_interaction}. The structures of 2D/3D P-Net and R-Net are detailed in~\ref{method:cnn_structures}. In~\ref{method:crf}, we describe the implementation of our CRF-Net. Training details are described in~\ref{method:implementation}.

\subsection{User Interaction-based Geodesic Distance Maps}\label{method:user_interaction}
In our method, scribbles are provided by the user to refine an initial automatic segmentation obtained by P-Net. A scribble labels a set of pixels as the foreground or background. Interactions with the same label are converted into a distance map. In~\cite{Xu2016}, the Euclidean distance was used due to its simplicity. However, the Euclidean distance treats each direction equally and does not take the image context into account. In contrast, the geodesic distance helps to better differentiate neighboring pixels with different appearances, and improve label consistency in homogeneous regions~\cite{Criminisi2008}. GeoF~\cite{Kohli2013} uses the geodesic distance to encode variable dependencies in the feature space and it is combined with Random Forests for semantic segmentation. However, it is not designed to deal with user interactions. We propose to encode user interactions via geodesic distance transforms for CNN-based segmentation. 

Suppose $\mathcal S_f$ and $\mathcal S_b$ represent the set of pixels belonging to foreground scribbles and background scribbles,  respectively. Let $i$ be a pixel in an image $\mathbf{I}$, then the unsigned geodesic distance from $i$ to the scribble set $\mathcal S (\mathcal S \in \{\mathcal S_f, \mathcal S_b\})$ is:
\begin{align}
G(i,\mathcal S,\mathbf{I}) = \min_{j\in S}{D_{geo}(i,j,\mathbf{I})}
\label{eq:dis}
\end{align}
\begin{align}
D_{geo}(i, j,\mathbf{I})= \min_{p \in \mathcal{P}_{i,j}}
{\int_0^1 \|\nabla \mathbf{I}(p(s)) \cdot \mathbf{u}(s)\| ds}
\label{eq:geodis}
\end{align}
 where $\mathcal{P}_{i,j}$ is the set of all paths between pixel $i$ and $j$. $p$ is one feasible path and it is parameterized by $s\in$ [0,1]. $\mathbf{u}(s) = p'(s)/\|p'(s)\|$ is a unit vector that is tangent to the direction of the path. If no scribbles are drawn for either the foreground or background, the corresponding geodesic distance map is filled with random numbers.

Fig.~\ref{fig:scribble_distance} shows an example of geodesic distance transforms of user interactions. The geodesic distance maps of user interactions and the initial automatic segmentation have the same size as $\mathbf{I}$. They are concatenated with the raw channels of $\mathbf{I}$ so that a concatenated image with $C_I$+3 channels is obtained, which is used as the input of the refinement network R-Net.




\subsection{Resolution-Preserving CNNs using Dilated Convolution}\label{method:cnn_structures}
CNNs in our method are designed to capture high-level features from a large receptive field without the loss of resolution of the feature maps. They are adapted from VGG-16~\cite{Simonyan2015} and made resolution-preserving. Fig.~\ref{fig:p-net} shows the structure of 2D and 3D P-Net. In 2D P-Net, the first 13 convolution layers are grouped into five blocks. The first and second blocks have two convolution layers respectively, and each of the remaining blocks has three convolution layers. The size of the convolution kernel is fixed as 3$\times$3 in all these convolution layers.  2D R-Net uses the same structure as 2D P-Net except that its number of input channels is $C_I$+3 and it employs user interactions in the CRF-Net. To obtain an exponential increase of the receptive field, VGG-16 uses a max-pooling and downsampling layer after each block. However, this implementation would decrease the resolution of feature maps exponentially. Therefore, to preserve resolution through the network, we remove the max-pooling and downsampling layers and use dilated convolution in each block. 

Let $\mathbf{I}$ be a 2D image of size $W\times H$, and let $K_{rq}$ be a square dilated convolution kernel with a size of (2$r$+1)$\times$(2$r$+1) and a dilation parameter $q$, where $r\in \mathbb{Z}$ and $q\in \mathbb{Z}$. The dilated convolution of $\mathbf{I}$ with $K_{rq}$ is defined as:
\begin{align}
\mathbf{I}_c(x,y)= \sum_{i=-r}^{r}\sum_{j=-r}^{r}\mathbf{I}(x-qi, y-qj)K_{rq}(i+r,j+r) 
\label{eq:dilated_convolution}
\end{align}
For 2D P-Net/R-Net, we set $r$ to 1 for block 1 to block 5, so the size of a convolution kernel becomes 3$\times$3. The dilation parameter in block $i$ is set to:
\begin{align}
q_i= d\times2^{i-1},  i=1,2,..., 5
\label{eq:dilated_factor}
\end{align}
 where $d \in \mathbb{Z}$ is a system parameter controlling the base dilation parameter of the network. We set $d$=1 in experiments.

The receptive field of a dilated convolution kernel $K_{rq}$ is (2$rq$+1)$\times$(2$rq$+1). Let $R_i\times R_i$ denote the receptive field of block $i$.  $R_i$ can be computed as:  
\begin{align}
	R_i = 2\Big(\sum_{j=1}^{i}{\tau_j \times (rq_j)\Big)}+1, i=1,2,...,5
	\label{eq:receptive_field}
\end{align}
where $\tau_j$ is the number of convolution layers in block $j$, with a value of 2, 2, 3, 3, 3 for the five blocks respectively.  
When $r$=1, the receptive field size of each block is $R_1$=4$d$+1, $R_2$=12$d$+1, $R_3$=36$d$+1, $R_4$=84$d$+1, $R_5$=180$d$+1, respectively. Thus, these blocks capture features at different scales. 

The stride of each convolution layer is set to 1. The number of output channels of convolution in each block is set to a fixed number $C$. In order to use multi-scale features, we concatenate the features from different blocks to get a composed feature of length 5$C$. This feature is fed into a classifier that is implemented by two additional layers as shown  in block 6 in Fig.~\ref{fig:p-net}(a). These two layers use  convolution kernels with size of 1$\times$1 and dilation parameter of 0. Block 6 gives each pixel an initial score of belonging to the foreground or background class. In order to get a more spatially consistent segmentation and add hard constraints when scribbles are given, we apply a CRF on the basis of the output from block 6. The CRF is implemented by a recurrent neural network (CRF-Net, detailed in~\ref{method:crf}), which can be jointly trained with P-Net or R-Net. The CRF-Net gives a regularized prediction for each pixel, which is fed into a cross entropy loss function layer.

Similar network structures are used by 3D P-Net/R-Net for 3D segmentation, as shown in Fig.~\ref{fig:p-net}(b). To reduce the memory consumption for 3D images, we use one downsampling layer before the resolution-preserving layers and compress the output features of block 1 to 5 by a factor four via 1$\times$1$\times$1 convolutions before the concatenation layer. 

\subsection{Back-propagatable CRF-Net with Freeform Pairwise Potentials and User Constraints}\label{method:crf}
In~\cite{Zheng2015a}, a CRF based on RNN was proposed and it can be trained by back-propagation. Rather than using Gaussian functions, we extend this CRF so that the pairwise potentials can be freeform functions and we refer to it as CRF-Net(f). In addition, we integrate user interactions in our CRF-Net(f) in the interactive refinement context, which is referred to as CRF-Net(fu). The CRF-Net(f) is connected to P-Net and the CRF-Net(fu) is connected to R-Net.

\begin{figure}[t]
	\centering
	\includegraphics[width=0.8\linewidth]{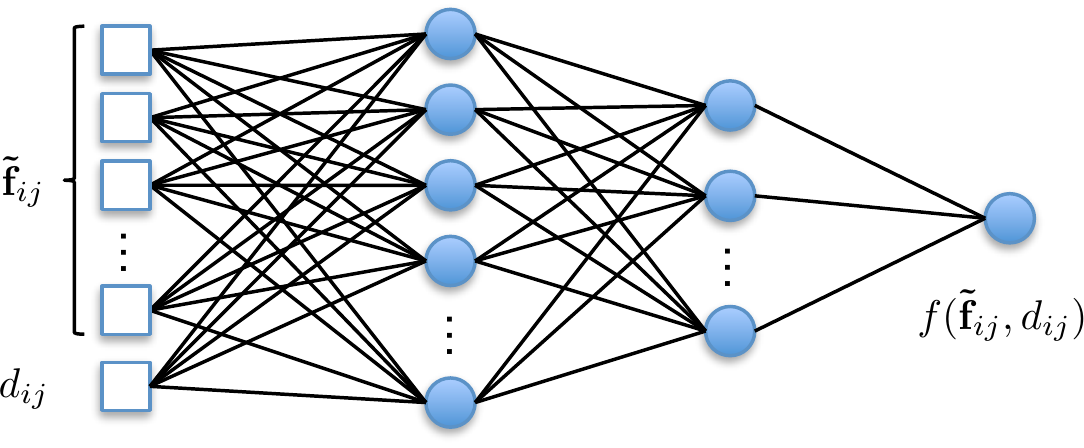}
	\caption[The Pairwise-Net for pairwise potential function]{ 
	The Pairwise-Net for pairwise potential function $f(\mathbf{\tilde{f}}_{ij}, d_{ij})$. $\mathbf{\tilde{f}}_{ij}$ is the difference of features between a pixel pair $i$ and $j$. $d_{ij}$ is the Euclidean distance between them. 
	} 
	\label{fig:pairwise-net}
\end{figure}

Let $\bf{X}$ be the label map assigned to an image $\mathbf{I}$ with a label set $\mathcal{L}$ = \{0, 1, ..., $L$ - 1\}. The Gibbs distribution 
$P(\mathbf{X}=\mathbf{x}|\mathbf{I}) = \frac{1}{Z(\mathbf{I})}\text{exp}(-E(\mathbf{x}|\mathbf{I}))$ models the probability of $\bf{X}$ given $\mathbf{I}$ in a CRF, where $Z(\mathbf{I})$ is the normalization factor known as the partition function, and $E(\mathbf{x})$ is the Gibbs energy:
\begin{align}
E(\mathbf{x}) = \sum_{i}\psi_u(x_i) + \sum_{(i,j)\in \mathcal N}\psi_p(x_i, x_j)
\label{eq:crf_energy}
\end{align}
where the unary potential $\psi_u(x_i)$ measures the cost of assigning label $x_i$ to pixel $i$, and the pairwise potential $\psi_p(x_i, x_j)$ is the cost of assigning labels $x_i, x_j$ to a pixel pair $i, j$. $\mathcal N$ is the set of all pixel pairs. In our method, the unary potential is obtained from P-Net or R-Net that gives classification scores at each pixel. The pairwise potential is:
\begin{align}
\psi_p(x_i, x_j)= \mu(x_i,x_j)f(\mathbf{\tilde{f}}_{ij}, d_{ij})
\label{eq:pairwise_potential}
\end{align}
where $d_{ij}$ is the Euclidean distance between pixels $i$ and $j$. $\mu(x_i,x_j)$ is the compatibility between the label of $i$ and that of $j$ represented by a matrix of size $L\times L$.  $\mathbf{\tilde{f}}_{ij} = \mathbf{f}_i - \mathbf{f}_j$, where $\mathbf{f}_i$ and $\mathbf{f}_j$ represent the feature vectors of $i$ and $j$, respectively.  The feature vectors can either be learned by a network or be derived from image features such as spatial location with intensity values. For experiments we used the latter one, as in~\cite{Zheng2015a, Krahenbuhl2011, Boykov2001} for simplicity and efficiency. $f$($\cdot$) is a function in terms of $\mathbf{\tilde{f}}_{ij}$ and $d_{ij}$. Instead of defining $f$($\cdot$) as a single Gaussian function~\cite{Boykov2001} or a combination of several Gaussian functions~\cite{Zheng2015a, Krahenbuhl2011}, we set it as a freeform function represented by a fully connected neural network (Pairwise-Net) which can be learned during training. The structure of Pairwise-Net is shown in Fig.~\ref{fig:pairwise-net}. The input is a vector composed of $\mathbf{\tilde{f}}_{ij}$ and $d_{ij}$. There are two hidden layers and one output layer.

Graph Cuts~\cite{Boykov2001, Szummer2008} can be used to minimize Eq.~\eqref{eq:crf_energy} when $\psi_p$($\cdot$) is submodular~\cite{Kolmogorov2004} such as when the segmentation is binary with $\mu$($\cdot$) being the delta function and $f$($\cdot$) being positive. However, this is not the case for our method since we learn $\mu$($\cdot$) and $f$($\cdot$) where $\mu$($\cdot$) may not be the delta function and $f$($\cdot$) could be negative. Continuous max-flow~\cite{Yuan2010} can also be used for the minimization, but its parameters are manually designed. Alternatively, mean-field approximation~\cite{Zheng2015a, Krahenbuhl2011, Vemulapalli2016} is often used to efficiently minimize the energy while allowing learning parameters by back-propagation. Instead of computing $P(\mathbf{X}|\mathbf{I})$ directly, an approximate distribution $Q(\mathbf{X}|\mathbf{I}) = \prod_iQ_i(x_i|\mathbf{I})$ is computed so that the KL-divergence $\mathbf{D}(Q||P)$ is minimized. This yields an iterative update of $Q_i(x_i|\mathbf{I})$ \cite{Zheng2015a, Krahenbuhl2011, Vemulapalli2016}. 
\begin{align}
Q_i(x_i|\mathbf{I})=\frac{1}{Z_i}e ^{-E(x_i)}=
\frac{1}{Z_i}e^{
-\psi_u(x_i)
-\phi_p(x_i)}\label{eq:update_rule}
\end{align}
\begin{align}
\phi_p(x_i=l|\mathbf{I})=
\sum_{l'\in \mathcal L}\mu(l,l')\sum_{(i,j)\in \mathcal N}f(\mathbf{\tilde{f}}_{ij}, d_{ij})Q_j(l'|\mathbf{I})   
\label{eq:update_rule2}
\end{align}
where $\mathcal L$ is the label set. $i$ and $j$ are a pixel pair. For the proposed CRF-Net(fu), with the set of user-provided scribbles $\mathcal S_{fb} = \mathcal S_f \cup \mathcal S_b$, we force the probability of pixels in the scribble set to be 1 or 0. The following equation is used as the update rule for each iteration:   

\begin{align}
Q_i(x_i|\mathbf{I})= 
\begin{cases}
1       & \quad \text{if } i \in \mathcal S_{fb}  \text{ and } x_i=s_i\\
0       & \quad \text{if } i \in \mathcal S_{fb}  \text{ and } x_i\neq s_i\\
\frac{1}{Z_i}e ^{-E(x_i)}  & \quad \text{otherwise } \\
\end{cases}\label{eq:update_rule_with_scrible}
\end{align}
where $s_i$ denotes the user-provided label of a pixel $i$ that is in the scribble set $\mathcal S_{fb}$. 
We follow the implementation in \cite{Zheng2015a} to update $Q$ through a multi-stage mean-field method in an RNN. Each mean-field layer splits Eq.~\eqref{eq:update_rule} into four steps including message passing, compatibility transform, adding unary potentials and normalizing \cite{Zheng2015a}.
\subsection{Implementation Details}\label{method:implementation}
The raster-scan algorithm~\cite{Criminisi2008} was used to compute geodesic distance transforms by applying a forward pass scanning and a backward pass scanning with a 3$\times$3 kernel for 2D and a 3$\times$3$\times$3 kernel for 3D.
It is fast due to accessing the image memory in contiguous blocks.
For the proposed CRF-Net with freeform pairwise potentials, two observations motivate us to use pixel connections based on local patches instead of full connections within the entire image. First, the permutohedral lattice implementation~\cite{Krahenbuhl2011, Zheng2015a} allows efficient computation of fully connected CRFs only when pairwise potentials are Gaussian functions. However, a method that relaxes the requirement of pairwise potentials as freeform functions represented by a network (Fig.~\ref{fig:pairwise-net}) cannot use that implementation and therefore would be inefficient for fully connected CRFs. Suppose an image with size $M\times N$, a fully connected CRF has $MN$($MN$-1) pixel pairs. For a small image with $M$=$N$=100, the number of pixel pairs would be almost 10$^8$, which requires not only a huge amount of memory but also long computational time. Second, though long-distance dependency helps to improve segmentation in most RGB images~\cite{Krahenbuhl2011, Zheng2015a, Chen2015iclr}, this would be very challenging for medical images since the contrast between the target and background is often low~\cite{Han2011ipmi}. In such cases, long-distance dependency may lead the label of a target pixel to be corrupted by the large number of background pixels with similar appearances. Therefore, to maintain a good efficiency and avoid long-distance corruptions, we define the pairwise connections for one pixel within a local patch centered on that. In our experiments, the patch size is set to 7$\times$7 for 2D images and 5$\times$5$\times$3 for 3D images.

We initialize $\mu$($\cdot$) as $\mu$($x_i$, $x_j$) = [$x_i\neq x_j$], where [$\cdot$] is the Iverson Bracket~\cite{Zheng2015a}.
A fully connected neural network (Pairwise-Net) with two hidden layers is used to learn the freeform pairwise potential function (Fig.~\ref{fig:pairwise-net}). The first and second hidden layers have 32 and 16 neurons, respectively. In practice, this network is implemented by an equivalent fully convolutional neural network with 1$\times$1$\times$1 kernels. We use a pre-training step to initialize the Pairwise-Net with an approximation of a contrast sensitive function~\cite{Boykov2001}:
\begin{align}
	f_0(\mathbf{\tilde{f}}_{ij},d_{ij})= \text{exp}\left(-\frac{||\mathbf{\tilde{f}}_{ij}||^2}{2\sigma^2\cdot F}\right)\cdot \frac{\omega}{d_{ij}}
	\label{eq:pairwise_potential_initialize}
\end{align}
where $F$ is the dimension of the feature vectors $\mathbf{f}_i$ and $\mathbf{f}_j$, and $\omega$ and $\sigma$ are two parameters controlling the magnitude and shape of the initial pairwise function respectively. In this initialization step, we set $\sigma$ to 0.08 and $\omega$ to 0.5 based on experience. Similar to~\cite{Krahenbuhl2011, Chen2016deeplab, Zheng2015a}, we set $\mathbf{f}_i$ and $\mathbf{f}_j$ as values in input channels (i.e, image intensity in our case) of P-Net for simplicity of implementation and for obtaining contrast-sensitive pairwise potentials. To pre-train the Pairwise-Net we generate a training set $T'=\{X',Y'\}$ with 100k samples, where $X'$ is the set of features simulating the concatenated $\mathbf{\tilde{f}}_{ij}$ and $d_{ij}$, and $Y'$ is the set of prediction values simulating $f_0(\mathbf{\tilde{f}}_{ij}, d_{ij})$. For each sample $s$ in $T'$, the feature vector $x'_s$ has a dimension of $F$+1 where the first $F$ dimensions represent the value of $\mathbf{\tilde{f}}_{ij}$ and the last dimension denotes $d_{ij}$. The $c$-th channel of $x'_s$ is filled with a random number $k'$, where $k'\sim Norm$(0, 2) for $c\leq F$ and  $k'\sim U$(0, 8) for $c =F$+1. The ground truth of prediction value $y'_s$ for $x'_s$ is obtained by Eq.~\eqref{eq:pairwise_potential_initialize}. After generating $X'$ and $Y'$, we use a Stochastic Gradient Descent (SGD) algorithm with a quadratic loss function to pre-train the Pairwise-Net.

For pre-processing, all the images are normalized by the mean value and standard variation of the training set. We apply data augmentation by vertical or horizontal flipping, random rotation with angle range [-$\pi$/8, $\pi$/8] and random zoom with scaling factor range [0.8, 1.25]. We use the cross entropy loss function and SGD algorithm for optimization with minibatch size 1, momentum 0.99 and weight decay 5$\times$10$^{-4}$. The learning rate is halved every 5k iterations. Since a proper initialization of P-Net and CRF-Net(f) is helpful for a faster convergence of the joint training, we train the P-Net with CRF-Net(f) in three steps. First, the P-Net is pre-trained with initial learning rate 10$^{-3}$ and maximal number of iterations 100k. Second, the Pairwise-Net in the CRF-Net(f) is pre-trained as described above. Third, the P-Net and CRF-Net(f) are jointly trained with initial learning rate 10$^{-6}$ and maximal number of iterations 50k.

\begin{figure}[t]
	\centering 
	\subfigure[]{\includegraphics[height=0.3\linewidth]{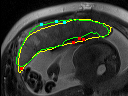}}
	\subfigure[]{\includegraphics[height=0.3\linewidth]{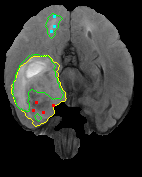}}
	\subfigure[]{\includegraphics[height=0.3\linewidth]{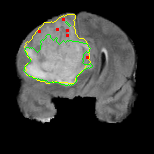}}
	\caption{
		Simulated user interactions on training images for placenta (a) and brain tumor (b, c). Green: automatic segmentation given by P-Net with CRF-Net(f). Yellow: ground truth. Red(cyan): simulated clicks on under-segmentation(over-segmentation).} 
	\label{fig:simulate_interaction}
\end{figure}

After the training of P-Net with CRF-Net(f), we automatically simulate user interactions to train R-Net with CRF-Net(fu). First, P-Net with CRF-Net(f) is used to obtain an automatic segmentation for each training image. It is compared with the ground truth to find mis-segmented regions. Then the user interactions on each mis-segmented region are simulated by randomly sampling $n$ pixels in that region. Suppose the size of one connected under-segmented or over-segmented  region is $N_m$, we set $n$ for that region to 0 if $N_m<$ 30 and $\lceil N_m$/100 $\rceil$ otherwise based on experience. 
Examples of simulated user interactions on training images are shown in Fig.~\ref{fig:simulate_interaction}. With these simulated user interactions on the initial segmentation of training data, the training of R-Net with CRF-Net(fu) is implemented through SGD, which is similar to the training of P-Net with CRF-Net(f). 

We implemented our 2D networks by Caffe\footnote{\url{http://caffe.berkeleyvision.org}}~\cite{Jia2014} and 3D networks by Tensorflow\footnote{\url{https://www.tensorflow.org}}~\cite{Abadi2016} using NiftyNet\footnote{\url{http://niftynet.io}}~\cite{Li2017}. Our training process was done via two 8-core E5-2623v3 Intel Haswells and two K80 NVIDIA GPUs and 128GB memory. The testing process with user interactions was performed on a MacBook Pro (OS X 10.9.5) with 16GB RAM and an Intel Core i7 CPU running at 2.5GHz and an NVIDIA
GeForce GT 750M GPU. A Matlab and PyQt GUI were developed for 2D and 3D interactive segmentation tasks, respectively.
(See supplementary videos)
\section{Experiments}
\subsection{Comparison Methods and Evaluation Metrics}
We compared our P-Net with FCN~\cite{Long2014} and DeepLab~\cite{Chen2016deeplab} for 2D segmentation and DeepMedic~\cite{Kamnitsas2017} and HighRes3DNet~\cite{Li2017} for 3D segmentation. Pre-trained models of FCN\footnote{\url{https://github.com/shelhamer/fcn.berkeleyvision.org}} and DeepLab\footnote{\url{https://bitbucket.org/deeplab/deeplab-public}} based on ImageNet were fine-tuned for 2D placenta segmentation. 
Since the input of FCN and DeepLab should have three channels, we duplicated each of the gray-level images twice and concatenated them into a three-channel image as the input. DeepMedic and HighRes3DNet were originally designed for multi-modality or multi-class 3D segmentation. We adapted them for single modality binary segmentation. We also compared 2D/3D P-Net with 2D/3D P-Net(b5) that only uses the features from block 5 (Fig.~\ref{fig:p-net}) instead of the concatenated multi-scale features. 

The proposed CRF-Net(f) with freeform pairwise potentials was compared with: 1). Dense CRF as an independent post-processing step for the  output of P-Net. We followed the implementation  in~\cite{Krahenbuhl2011, Chen2016deeplab, Kamnitsas2017}.
The parameters of this CRF were manually tuned based on a coarse-to-fine search scheme as suggested by~\cite{Chen2016deeplab}, and 2). CRF-Net(g) which refers to the CRF that can be trained jointly with CNNs by using Gaussian pairwise potentials~\cite{Zheng2015a}.

We compared three methods to deal with user interactions. 1). Min-cut user-editing~\cite{Rother2004}, where the initial probability map (output of P-Net in our case) is combined with user interactions to solve an energy minimization problem with min-cut~\cite{Boykov2001}; 2). Using the Euclidean distance of user interactions in R-Net, which is referred to as R-Net(Euc), and 3). The proposed R-Net with the geodesic distance of user interactions.  

We also compared DeepIGeoS with several other interactive segmentation methods. For 2D slices, DeepIGeoS was compared with: 1). Geodesic Framework~\cite{Bai2007} that computes a probability based on the geodesic distance from user-provided scribbles for pixel classification; 2). Graph Cuts~\cite{Boykov2001} that models segmentation as a min-cut problem based on user interactions; 3). Random Walks~\cite{Grady2006a} that assigns a pixel with a label based on the probability that a random walker reaches a foreground or background seed first, and 4). SlicSeg~\cite{Wang2016} that uses Online Random Forests to learn from the scribbles and predict the remaining pixels. For 3D images, DeepIGeoS was compared with GeoS~\cite{Criminisi2008} and ITK-SNAP~\cite{Yushkevich2006}. Two users (an Obstetrician and a Radiologist) 
respectively used these interactive methods to segment every test image until the result was visually acceptable.

For quantitative evaluation, we measured the Dice score and the average symmetric surface distance (ASSD).  
\begin{align}
\text{Dice}=\frac{2|\mathcal{R}_a\cap \mathcal{R}_b|}{|\mathcal{R}_a|+|\mathcal{R}_b|}
\end{align} 
where $\mathcal{R}_a$ and $\mathcal{R}_b$ represent the region segmented by the algorithm and the ground truth, respectively.
\begin{align}
\text{ASSD}=\frac{1}{|\mathcal{S}_a|+|\mathcal{S}_b|}
\left(
\sum_{i\in \mathcal{S}_a}d(i,\mathcal{S}_b)+\sum_{i\in \mathcal{S}_b}
d(i,\mathcal{S}_a)
\right)
\label{eq:assd}
\end{align} 
where $\mathcal{S}_a$ and $\mathcal{S}_b$ represent the set of surface points of the target segmented by the algorithm and the ground truth, respectively. $d(i,\mathcal{S}_b)$ is the shortest Euclidean distance between $i$ and $\mathcal{S}_b$. We used the Student's $t$-test to compute the $p$-value in order to see whether the results of two algorithms significantly differ from each other. 

\subsection{2D Placenta Segmentation from Fetal MRI}
\subsubsection{Clinical Background and Experiments Setting}
Fetal MRI is an emerging diagnostic tool complementary to ultrasound due to its large field of view and good soft tissue contrast. Segmenting the placenta from fetal MRI is important for fetal surgical planning such as in the case of twin-to-twin transfusion syndrome~\cite{Deprest2010}. Clinical fetal MRI data are often acquired with a large slice thickness for good contrast-to-noise ratio. Movement of the fetus can lead to inhomogeneous appearances between slices. In addition, the location and orientation of the placenta vary largely between individuals. These factors make automatic and 3D segmentation of the placenta a challenging task~\cite{Alansary2016}. Interactive 2D slice-based segmentation is expected to achieve more robust results~\cite{Wang2016, Wang2016a}. The 2D segmentation results can also be used for motion correction and high-resolution volume reconstruction~\cite{Keraudren2014}. 

We collected clinical MRI scans for 25 pregnancies in the second trimester. The data were acquired in axial view with pixel size between 0.7422 mm$\times$0.7422 mm and 1.582 mm$\times$1.582 mm and slice thickness 3 - 4 mm. Each slice was resampled with a uniform pixel size of 1 mm$\times$1 mm and cropped by a box of size 172$\times$128 containing the placenta. 
We used 17 volumes with 624 slices for training, three volumes with 122 slices for validation and five volumes with 179 slices for testing. The ground truth was manually delineated by a experienced Radiologist. 

\subsubsection{Automatic Segmentation by 2D P-Net with CRF-Net(f)}
\begin{figure}[t]
	\centering
	\includegraphics[width=1.0\linewidth]{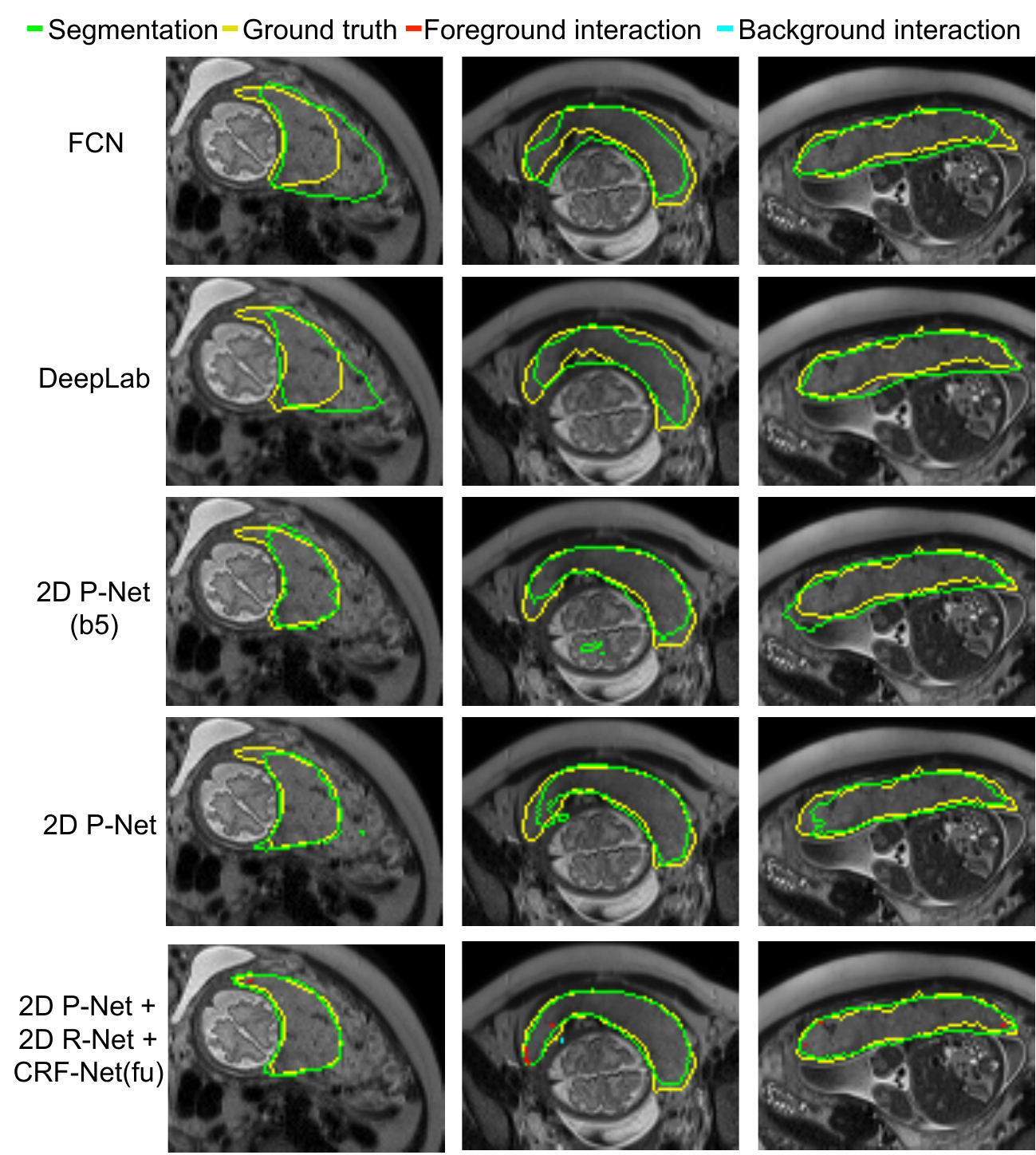}
	\caption[Initial automatic segmentation results of the placenta]{ 
		Initial automatic segmentation results of the placenta by 2D P-Net. 2D P-Net(b5) only uses the features from block 5 shown in Fig.~\ref{fig:p-net}(a) rather than the concatenated multi-scale features. The last row shows interactively refined results by DeepIGeoS.} 
	\label{fig:p-net_output}
\end{figure}

\begin{figure}[t]
	\centering
	\includegraphics[width=1.0\linewidth]{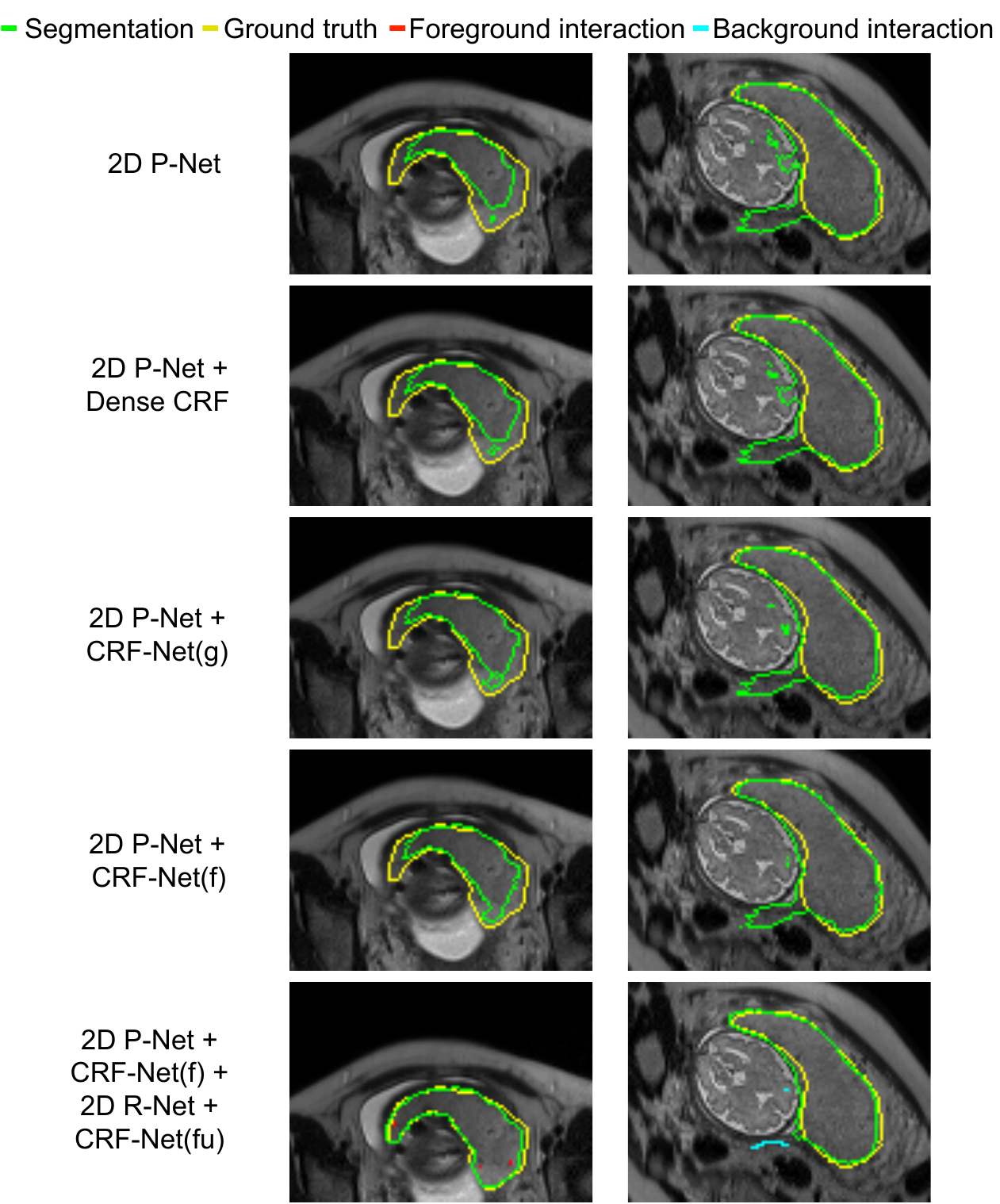}
	\caption[Visual comparison of placenta segmentation by 2D P-Net with different CRFs.]{ 
		Placenta segmentation by 2D P-Net with different CRFs. The last row shows interactively refined results by DeepIGeoS.
		} 
	\label{fig:p-net_crf}
\end{figure}

\begin{table}
	\centering
	\small
	\caption{Quantitative comparison of 2D placenta segmentation by different networks and CRFs. CRF-Net(g)~\cite{Zheng2015a} constrains pairwise potential as Gaussian functions. CRF-Net(f) is our proposed CRF that learns freeform pairwise potential functions. Significant improvement from 2D P-Net ($p$-value $<$ 0.05) is shown in bold font.}
	\label{lab:p-net_dice}
	\begin{tabular}{lll}
		
		\hline
		Method & Dice(\%) & ASSD(pixels) \\ \hline
		FCN~\cite{Long2014}       & 81.47$\pm$11.40 & 2.66$\pm$1.39 \\ 
		DeepLab~\cite{Chen2016deeplab}   & 83.38$\pm$9.53  & 2.20$\pm$0.84 \\
		2D P-Net(b5) & 83.16$\pm$13.01 & 2.36$\pm$1.66 \\
		2D P-Net     & 84.78$\pm$11.74 & 2.09$\pm$1.53 \\  
		2D P-Net + Dense CRF & 84.90$\pm$12.05          & 2.05$\pm$1.59 \\ 
		2D P-Net + CRF-Net(g)       & \textbf{85.44$\pm$12.50} & \textbf{1.98$\pm$1.46} \\ 
		2D P-Net + CRF-Net(f)       & \textbf{85.86$\pm$11.67} & \textbf{1.85$\pm$1.30} \\ \hline
	\end{tabular}
\end{table}
	
Fig.~\ref{fig:p-net_output} shows the automatic segmentation results obtained by different networks. It shows that FCN is able to capture the main region of the placenta. However, the segmentation results are blob-like with smooth boundaries. 
DeepLab is better than FCN, but its blob-like results are similar to those of FCN. This is mainly due to the downsampling and upsampling procedure employed by these methods. In contrast, 2D P-Net(b5) and 2D P-Net obtain more detailed results. 
It can be observed that 2D P-Net achieves better results than the other three networks. However, there are still some obvious mis-segmented regions by 2D P-Net. 
Table \ref{lab:p-net_dice} presents a quantitative comparison of these networks based on all the testing data. 2D P-Net achieves a Dice score of 84.78$\pm$11.74\% and an ASSD of 2.09$\pm$1.53 pixels, and it performs better than the other three networks. 

Based on 2D P-Net, we investigated the performance of different CRFs. A visual comparison between Dense CRF, CRF-Net(g) with Gaussian pairwise potentials and CRF-Net(f) with freeform pairwise potentials is shown in Fig.~\ref{fig:p-net_crf}. In the first column, the placenta is under-segmented by 2D P-Net. Dense CRF leads to very small improvements on the result. CRF-Net(g) and CRF-Net(f) improve the result by preserving more placenta regions, and the later shows a better segmentation. In the second column, 2D P-Net obtains an over-segmentation of adjacent fetal brain and maternal tissues. Dense CRF does not improve the segmentation noticeably, but CRF-Net(g) and CRF-Net(f) remove more over-segmentated areas. CRF-Net(f) shows a better performance than the other two CRFs. The quantitative evaluation of these three CRFs is presented in Table~\ref{lab:p-net_dice}, which shows Dense CRF leads to a result that is very close to that of 2D P-Net ($p$-value $>$ 0.05), while the last two CRFs significantly improve the segmentation ($p$-value $<$ 0.05). In addition, CRF-Net(f) is better than CRF-Net(g). 
Fig.~\ref{fig:p-net_crf} and Table~\ref{lab:p-net_dice} indicate that large mis-segmentation exists in some images, therefore we use 2D R-Net with CRF-Net(fu) to refine the segmentation interactively in the following.

\subsubsection{Interactive Refinement by 2D R-Net with CRF-Net(fu) }

\begin{figure*}[t]
	\centering
	\includegraphics[width=1.0\linewidth]{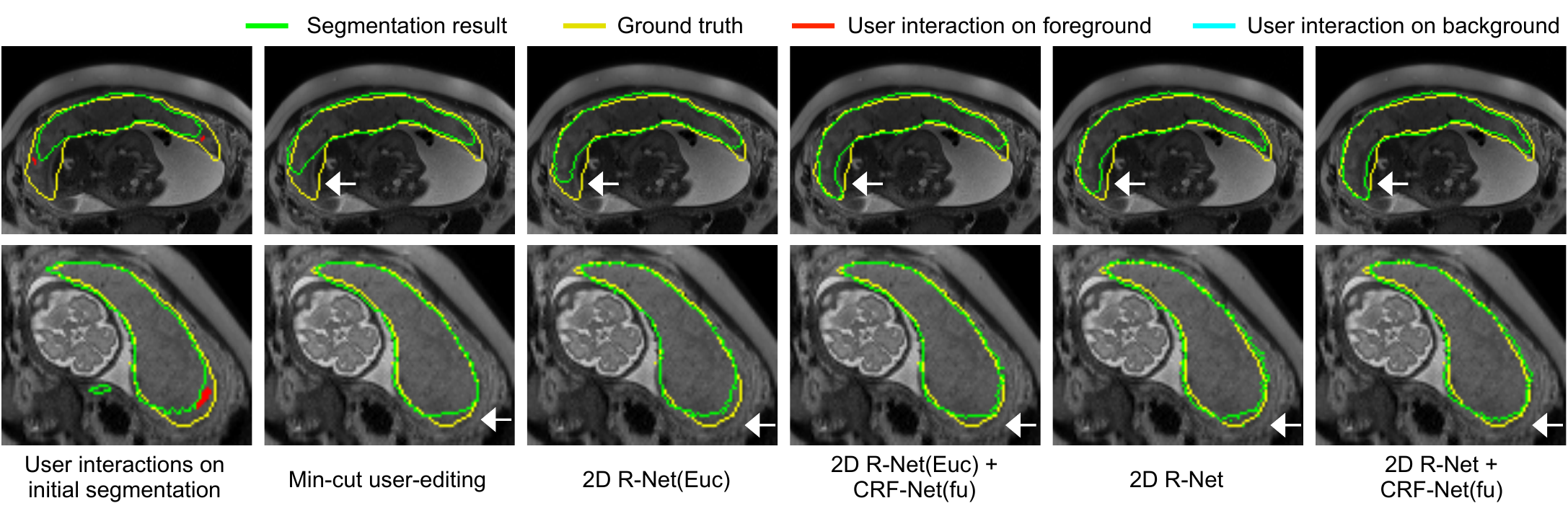}
	\caption[Visual comparison of different refinement methods for 2D placenta segmentation.]{ 
		Visual comparison of different refinement methods for 2D placenta segmentation. The first column shows the initial automatic segmentation obtained by 2D P-Net + CRF-Net(f), on which user interactions are added for refinement. The remaining columns show refined results. 2D R-Net(Euc) is a counterpart of the proposed 2D R-Net and it uses Euclidean distance. White arrows show the difference in local details.} 
	\label{fig:r_net_crf}
\end{figure*}

\begin{figure*}[t]
	\centering
	\includegraphics[width=0.9\linewidth]{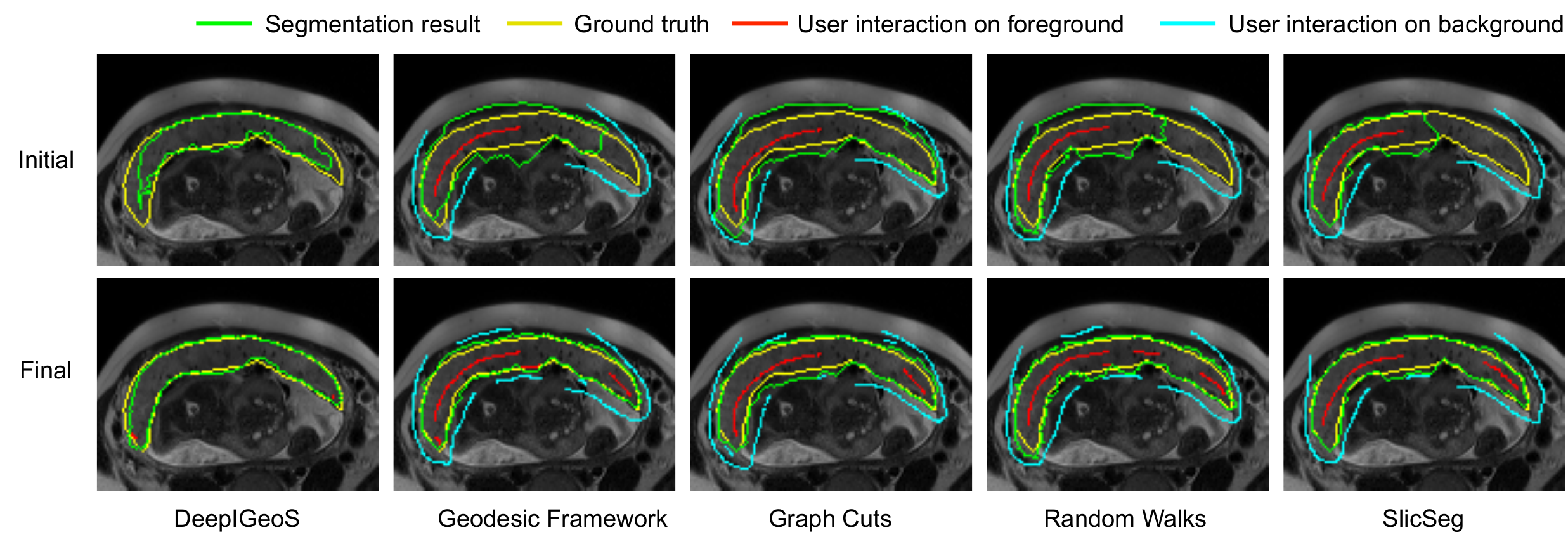}
	\caption[Visual comparison of DeepIGeoS and other interactive methods for 2D placenta segmentation.]{ 
		Visual comparison of DeepIGeoS and other interactive methods for 2D placenta segmentation. 
		The first row shows initial scribbles (except for DeepIGeoS) and the resulting segmentation. The second row shows final refined results with the entire set of scribbles. The user decided on the level of interaction required to achieve a visually acceptable result. } 
	\label{fig:compare_interactive_placenta}
\end{figure*}

\begin{table}
	\centering
	\small
	\caption{Quantitative evaluation of different refinement methods for 2D placenta segmentation. The initial segmentation is obtained by 2D P-Net + CRF-Net(f). 2D R-Net(Euc) uses Euclidean distance instead of geodesic distance. Significant improvement from 2D R-Net ($p$-value $<$ 0.05) is shown in bold font.}
	\label{lab:r-net_crf_dice}
	\begin{tabular}{lll}
		\hline
		Method & Dice(\%) & ASSD(pixels) \\ \hline
		Before refinement & 85.86$\pm$11.67 & 1.85$\pm$1.30 \\
		Min-cut user-editing & 87.04$\pm$9.79 & 1.63$\pm$1.15 \\
		2D R-Net(Euc)       & 88.26$\pm$10.61 & 1.54$\pm$1.18 \\ 
		
		2D R-Net       & 88.76$\pm$5.56  & 1.31$\pm$0.60 \\
		2D R-Net(Euc) + CRF-Net(fu) & 88.71$\pm$8.42  & 1.26$\pm$0.59 \\
		2D R-Net + CRF-Net(fu)  & \textbf{89.31$\pm$5.33}  & \textbf{1.22$\pm$0.55} \\  
		 \hline
	\end{tabular}
\end{table}
Fig.~\ref{fig:r_net_crf} shows examples of interactive refinement based on 2D R-Net with CRF-Net(fu) that uses freeform pairwise potentials and employs user interactions as hard constraints. The first column in Fig.~\ref{fig:r_net_crf} shows initial segmentation results obtained by 2D P-Net + CRF-Net(f). The user provides clicks/scribbles to indicate the foreground (red) or the background (cyan). 
The second to last column in Fig.~\ref{fig:r_net_crf} show the results for five variations of refinement. 
These refinement methods correct most of the mis-segmented areas but perform at different levels in dealing with local details, as indicated by white arrows. Fig.~\ref{fig:r_net_crf} shows 2D R-Net with geodesic distance performs better than min-cut user-editing and 2D R-Net(Euc) that uses Euclidean distance. CRF-Net(fu) can further improve the segmentation. For quantitative comparison, we measured the segmentation accuracy after the first iteration of user refinement (giving user interactions to mark all the main mis-segmented regions and applying refinement once), in which the same initial segmentation and the same set of user interactions were used by the five refinement methods. The results are presented in Table~\ref{lab:r-net_crf_dice}, which shows the combination of the proposed 2D R-Net using geodesic distance and CRF-Net(fu) leads to more accurate segmentations than the other refinement methods with the same set of user interactions. The Dice score and ASSD of 2D R-Net +  CRF-Net(fu) are 89.31$\pm$5.33\% and 1.22$\pm$0.55 pixels, respectively. 

\subsubsection{Comparison with Other Interactive Methods}

\begin{figure}[t]
	\centering
	\includegraphics[width=1.0\linewidth]{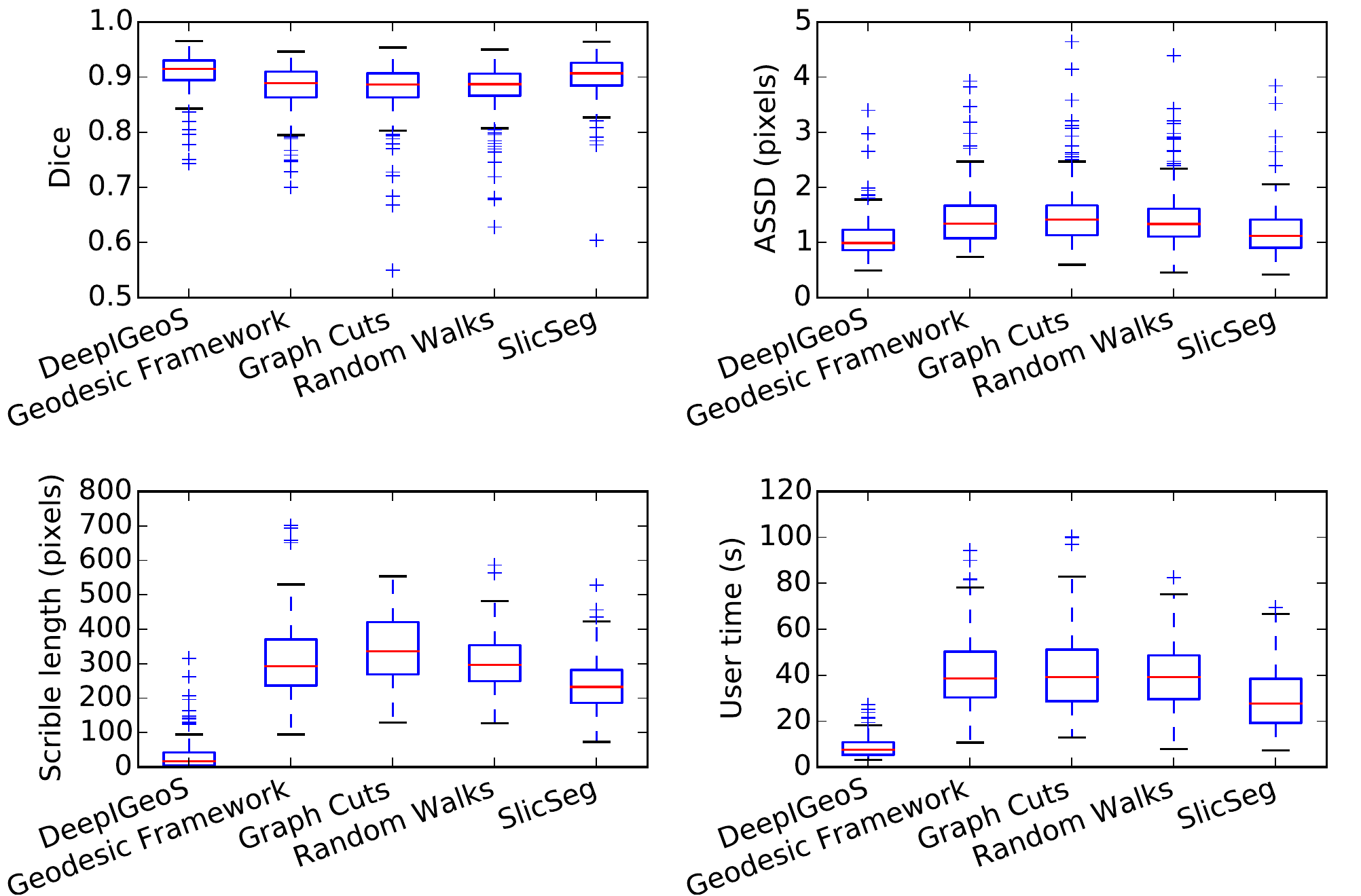}
	\caption[Quantitative comparison of 2D placenta segmentation by different interactive methods]{ 
		Quantitative comparison of 2D placenta segmentation by different interactive methods in terms of Dice, ASSD, total interactions (scribble length) and user time.} 
	\label{fig:placenta_compare_other}
\end{figure}
Fig.~\ref{fig:compare_interactive_placenta} shows a visual comparison between DeepIGeoS and Geodesic Framework~\cite{Bai2007}, Graph Cuts~\cite{Boykov2001}, Random Walks~\cite{Grady2006a} and SlicSeg~\cite{Wang2016} for 2D placenta segmentation. The first row shows the initial scribbles and the resulting segmentation. Notice no initial scribbles are needed for DeepIGeoS. The second row shows refined results, where DeepIGeoS only needs two short strokes to get an accurate segmentation, while the other methods require far more scribbles to get similar results. Quantitative comparison of these methods based on the final segmentation given by the two users is presented in Fig.~\ref{fig:placenta_compare_other}. It shows these methods achieve similar accuracy, but DeepIGeoS requires far fewer user interactions and less time. (See supplementary video 1)

\subsection{3D Brain Tumor Segmentation from FLAIR Images }
\subsubsection{Clinical Background and Experiments Setting}
Gliomas are the most common brain tumors in adults with little improvement in treatment effectiveness despite considerable research works~\cite{Menze2015a}. With the development of medical imaging, brain tumors can be imaged by different MR protocols with different contrasts. For example, T1-weighted images highlight enhancing part of the tumor and FLAIR acquisitions highlight the peritumoral edema. Segmentation of brain tumors can provide better volumetric measurements and therefore has enormous potential value for improved diagnosis, treatment planning, and follow-up of individual patients. However, automatic brain tumor segmentation remains technically challenging because 1) the size, shape, and localization of brain tumors have considerable variations among patients; 2) the boundaries between adjacent structures are often ambiguous. 

In this experiment, we investigate interactive segmentation of the whole tumor from FLAIR images. We used the 2015 Brain Tumor Segmentation Challenge (BraTS)~\cite{Menze2015a} training set with images of 274 cases. The ground truth were manually delineated by several experts. Differently from previous works using this dataset for multi-label and multi-modality segmentation~\cite{Kamnitsas2017,Fidon2017a}, as a first demonstration of deep interactive segmentation in 3D, we only use FLAIR images in the dataset and only segment the whole tumor. We randomly selected 234 cases for training and used the remaining 40 cases for testing. All these images had been skull-stripped and resampled to size of 240$\times$240$\times$155 with isotropic resolution 1mm$^3$. We cropped each image based on the bounding box of its non-zero region. The feature channel number of 3D P-Net and R-Net was $C = 16$. 
 
\subsubsection{Automatic Segmentation by 3D P-Net with CRF-Net(f)}

\begin{figure}[t]
	\centering
	\includegraphics[width=1.0\linewidth]{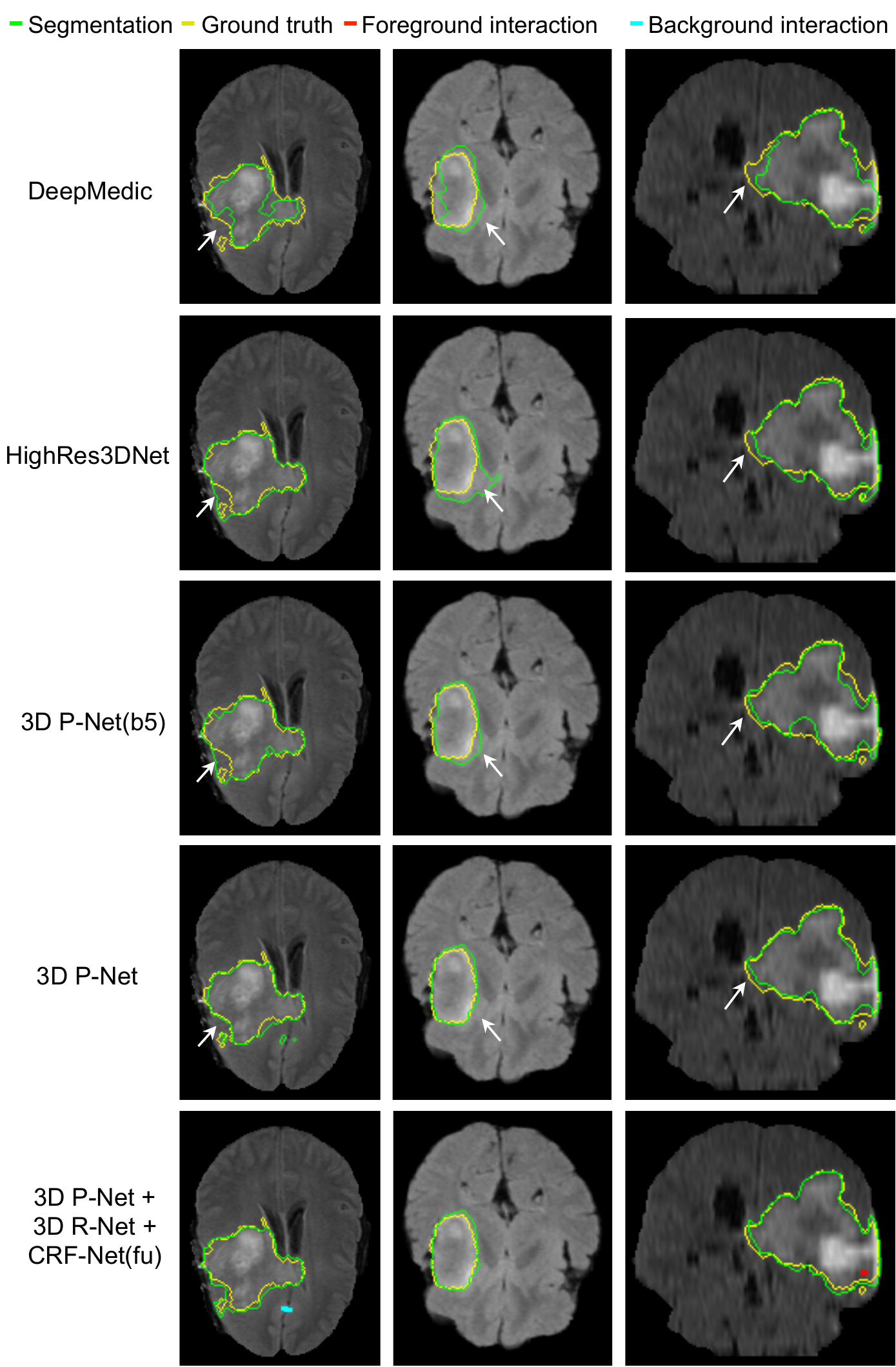}
	\caption[Initial automatic 3D segmentation of brain tumor]{ 
		Initial automatic 3D segmentation of brain tumor by different networks. 
		The last row shows interactively refined results by DeepIGeoS.
	} 
	\label{fig:tumor_p_net}
\end{figure}

\begin{figure}[t]
	\centering
	\includegraphics[width=1.0\linewidth]{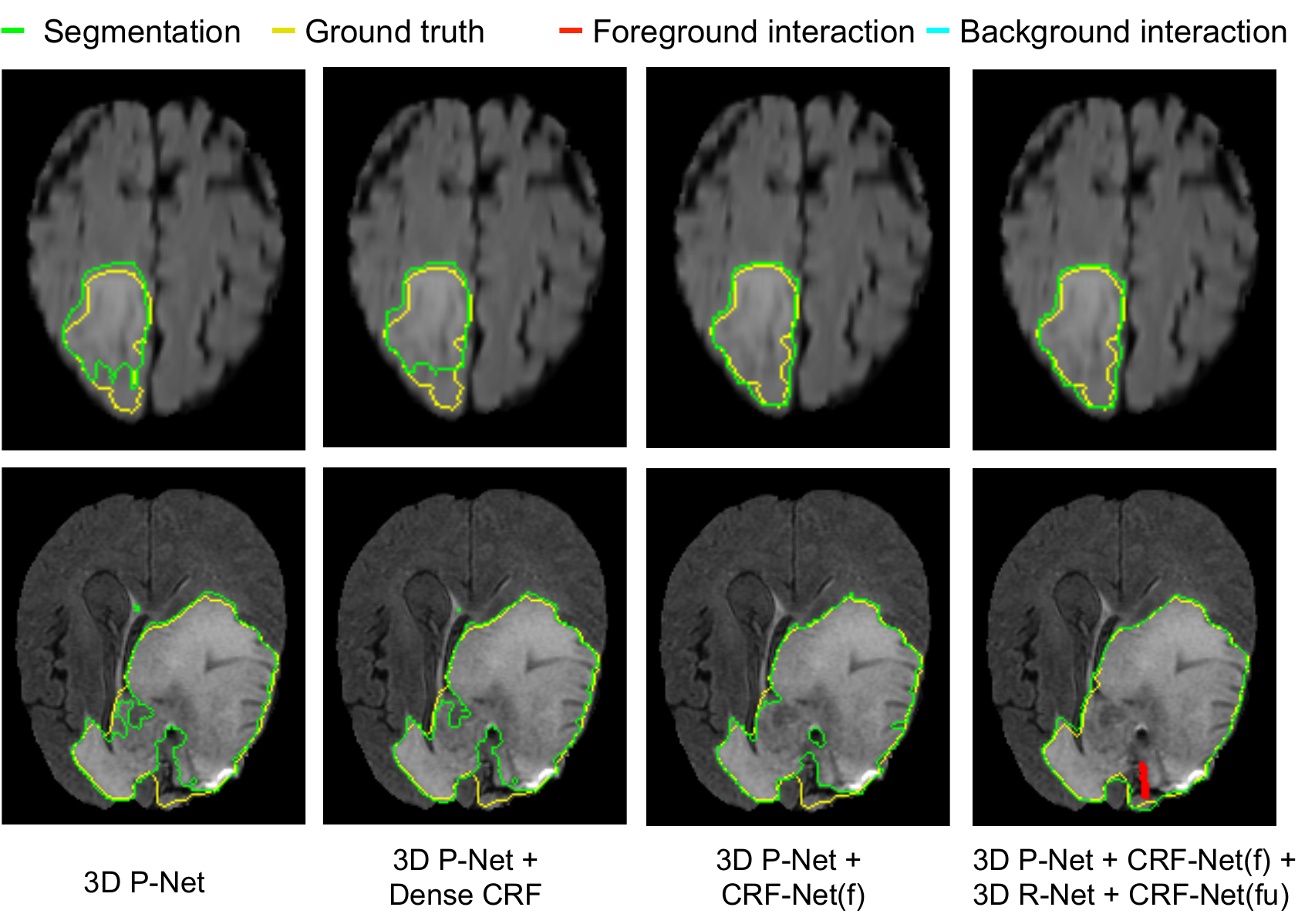}
	\caption[Visual comparison between Dense CRF and the proposed CRF-Net(f) for 3D brain tumor segmentation]{ 
		Visual comparison between Dense CRF and the proposed CRF-Net(f) for 3D brain tumor segmentation. 
		The last column shows interactively refined results by DeepIGeoS.
	} 
	\label{fig:tumor_p_net_crf}
\end{figure}

\begin{table}
	\centering\label{key}
	\small
	\caption{Quantitative comparison of 3D brain tumor segmentation by different networks and CRFs. 
		Significant improvement from 3D P-Net ($p$-value $<$ 0.05) is shown in bold font.}
	\label{lab:tumor_p_net}
	\begin{tabular}{lll}
		
		\hline
		Method    & Dice(\%) & ASSD(pixels) \\ \hline
		DeepMedic~\cite{Kamnitsas2017}       & 83.87$\pm$8.72 & 2.38$\pm$1.52 \\ 
		HighRes3DNet~\cite{Li2017}   & 85.47$\pm$8.66 & 2.20$\pm$2.24 \\
		3D P-Net(b5) & 85.36$\pm$7.34 & 2.21$\pm$2.13 \\
		3D P-Net     & 86.68$\pm$7.67 & 2.14$\pm$2.17 \\  
		3D P-Net + Dense CRF & 87.06$\pm$7.23 & 2.10$\pm$2.02 \\ 
		3D P-Net + CRF-Net(f) & \textbf{87.55$\pm$6.72} & \textbf{2.04$\pm$1.70} \\ \hline
	\end{tabular}
	
\end{table}
Fig.~\ref{fig:tumor_p_net} shows examples of automatic segmentation of brain tumor by 3D P-Net, which is compared with DeepMedic~\cite{Kamnitsas2017}, HighRes3DNet~\cite{Li2017} and 3D P-Net(b5). In the first column, DeepMedic segments the tumor roughly, with some missed regions near the boundary. HighRes3DNet reduces the missed regions but leads to some over-segmentation. 3D P-Net(b5) obtains a similar result to that of HighRes3DNet. In contrast, 3D P-Net achieves a more accurate segmentation, which is closer to the ground truth. More examples in the second and third column in Fig.~\ref{fig:tumor_p_net} also show 3D P-Net outperforms the other networks. Quantitative evaluation of these four networks is presented in Table~\ref{lab:tumor_p_net}.
DeepMedic achieves an average dice score of 83.87$\%$. HighRes3DNet and 3D P-Net(b5) achieve similar performance, and they are better than DeepMedic. 3D P-Net outperforms these three counterparts with
86.68$\pm$7.67\% in terms of Dice and 2.14$\pm$2.17 pixels in terms of ASSD. Note that the proposed 3D P-Net has far fewer parameters compared with HighRes3DNet. It is more memory efficient and therefore can perform inference on a 3D volume in interactive time.    

Since CRF-RNN~\cite{Zheng2015a} was only implemented for 2D, in the context of 3D segmentation we only compared 3D CRF-Net(f) with 3D Dense CRF~\cite{Kamnitsas2017} that uses manually tuned parameters. Visual comparison between these two types of CRFs working with 3D P-Net is shown in Fig.~\ref{fig:tumor_p_net_crf}. It can be observed that CRF-Net(f) achieves more noticeable improvement compared with Dense CRF that is used as post-processing without end-to-end learning. Quantitative measurement of Dense CRF and CRF-Net(f) is listed in Table~\ref{lab:tumor_p_net}. It shows that only CRF-Net(f) obtains significantly better segmentation than 3D P-Net with $p$-value $<$ 0.05. 
\subsubsection{Interactive Refinement by 3D R-Net with CRF-Net(fu) }

\begin{figure*}[t]
	\centering
	\includegraphics[width=0.9\linewidth]{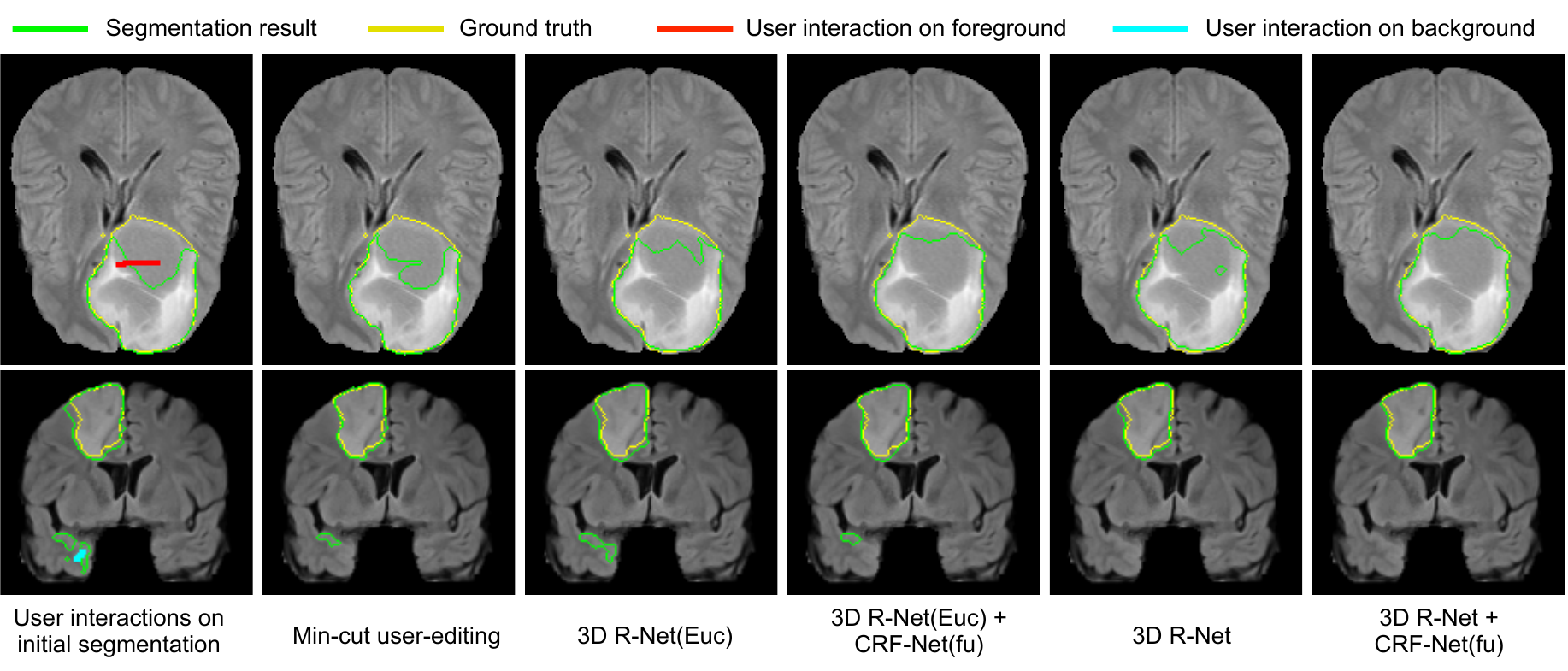}
	\caption[Visual comparison of different refinement methods for 3D brain tumor segmentation.]{ 
			Visual comparison of different refinement methods for 3D brain tumor segmentation. The initial segmentation is obtained by 3D P-Net + CRF-Net(f), on which user interactions are given. 3D R-Net(Euc) is a counterpart of the proposed 3D R-Net and it uses Euclidean distance. } 
	\label{fig:tumor_r_net_crf}
\end{figure*}

\begin{table}
	\centering
	\small
	\caption{Quantitative comparison of different refinement methods for 3D brain tumor segmentation with the same set of scribbles. The segmentation before refinement is obtained by 3D P-Net + CRF-Net(f). 3D R-Net(Euc) uses Euclidean distance instead of geodesic distance. Significant improvement from 3D R-Net ($p$-value $<$ 0.05) is shown in bold font.}
	\label{lab:tumor_r_net_crf}
	\begin{tabular}{lll}
		
		\hline
		Method & Dice(\%) & ASSD(pixels) \\ \hline
		Before refinement & 87.55$\pm$6.72 & 2.04$\pm$1.70 \\
		Min-cut user-editing & 88.41$\pm$7.05 & 1.74$\pm$1.53 \\
		3D R-Net(Euc) & 88.82$\pm$7.68 &  1.60$\pm$1.56 \\ 
		
		3D R-Net & 89.30$\pm$6.82 & 1.52$\pm$1.37 \\
		3D R-Net(Euc) + CRF-Net(fu) & 89.27$\pm$7.32 & 1.48$\pm$1.22 \\
		3D R-Net + CRF-Net(fu) & \textbf{89.93$\pm$6.49} & \textbf{1.43$\pm$1.16} \\  
		\hline
	\end{tabular}
	
\end{table}

Fig.~\ref{fig:tumor_r_net_crf} shows examples of interactive refinement of brain tumor segmentation using 3D R-Net with CRF-Net(fu). The initial segmentation is obtained by 3D P-Net + CRF-Net(f). With the same set of user interactions, we compared the refined results of min-cut user-editing and four variations of 3D R-Net: using geodesic or Euclidean distance transforms with or without CRF-Net(fu). Fig.~\ref{fig:tumor_r_net_crf} shows that min-cut user-editing achieves a small improvement. 
It can be found that more accurate results are obtained by using geodesic distance than using Euclidean distance, and CRF-Net(fu) can further help to improve the segmentation. For quantitative comparison, we measured the segmentation accuracy after the first iteration of refinement, in which the same set of scribbles were used for different refinement methods. The quantitative evaluation is listed in Table~\ref{lab:tumor_r_net_crf}, showing that the proposed 3D R-Net with geodesic distance and CRF-Net(fu) achieves higher accuracy than the other variations with a Dice score of 89.93$\pm$6.49\% and ASSD of 1.43$\pm$1.16 pixels.

\subsubsection{Comparison with Other Interactive Methods}
\begin{figure}[t]
	\centering
	\includegraphics[width=1.0\linewidth]{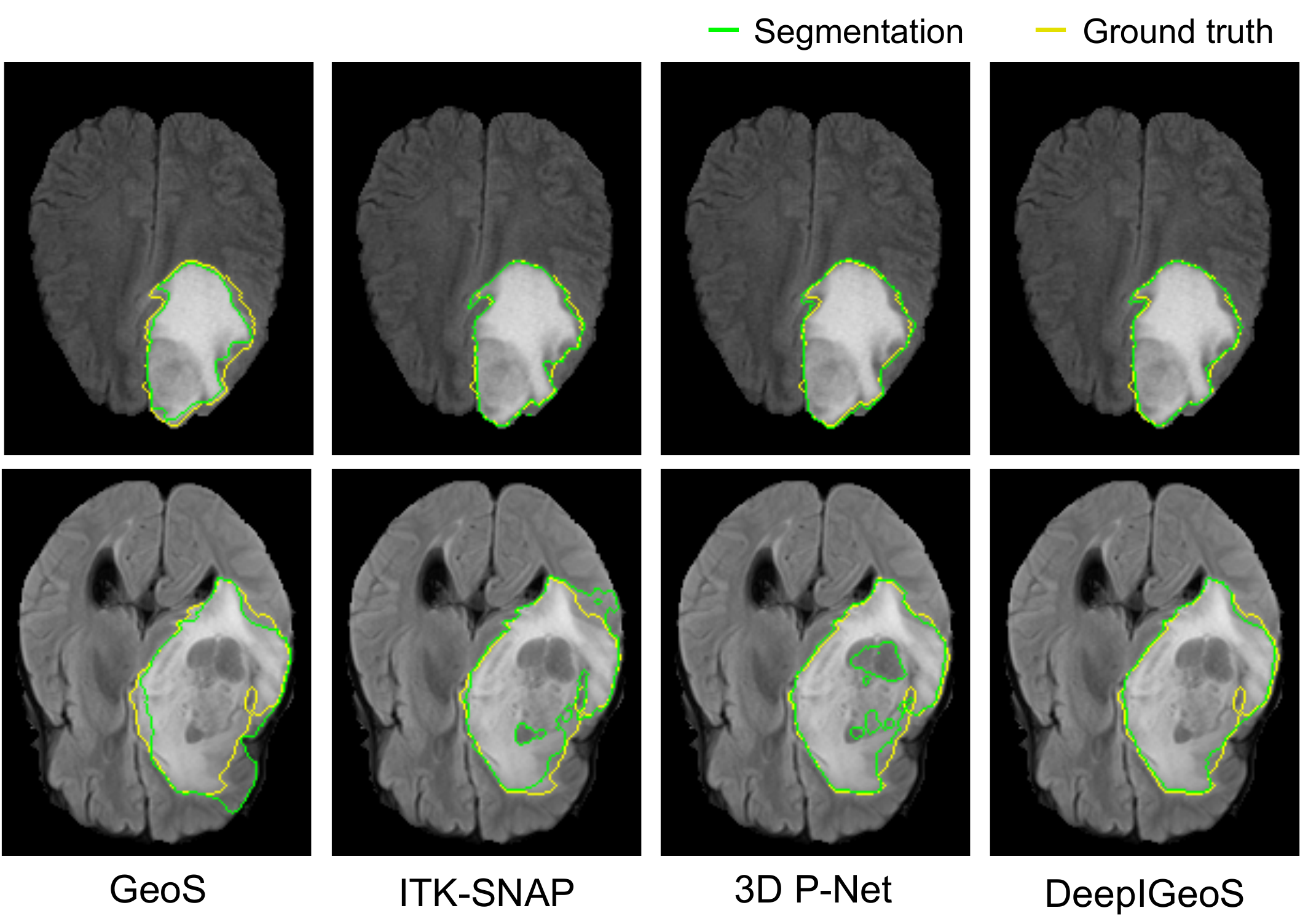}
	\caption[Visual comparison of 3D brain tumor segmentation]{ 
		Visual comparison of 3D brain tumor segmentation using GeoS, ITK-SNAP, and DeepIGeoS that is based on 3D P-Net.} 
	\label{fig:tumor_others}
\end{figure}
\begin{figure}[t]
	\centering
	\includegraphics[width=1.0\linewidth]{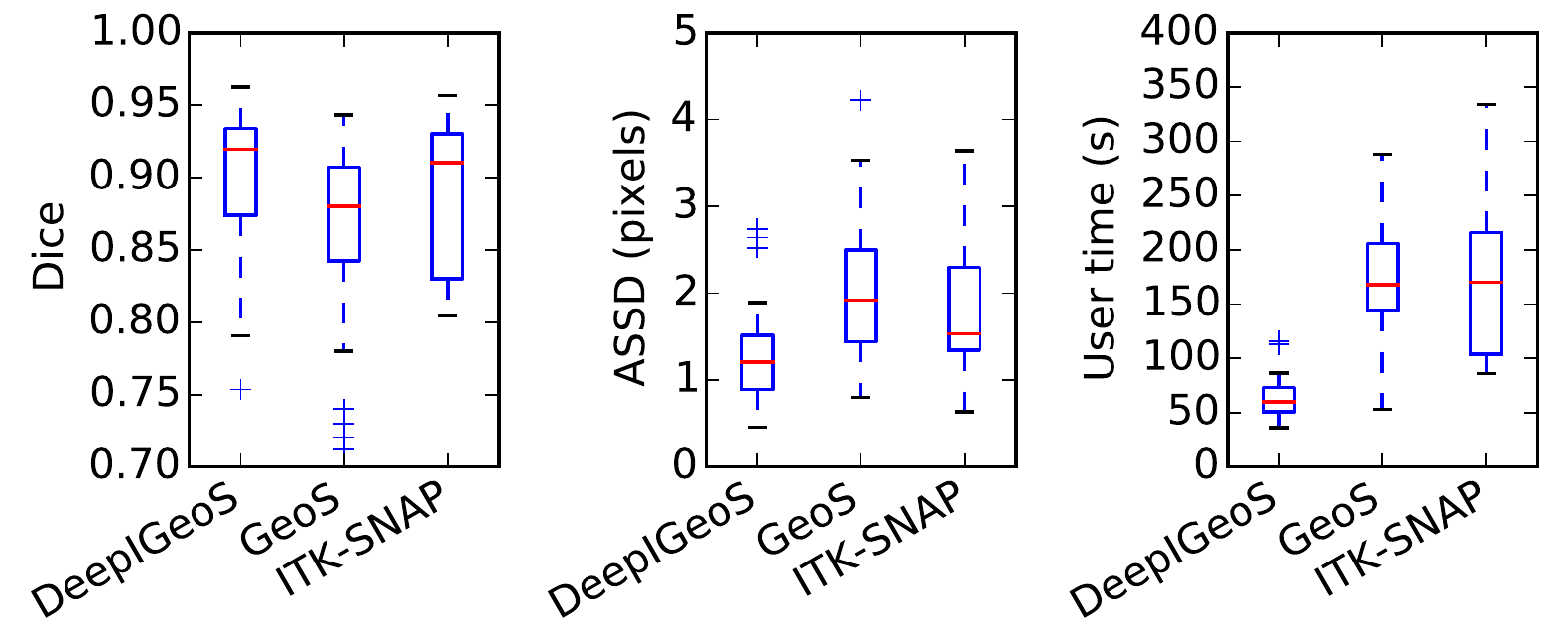}
	\caption[Quantitative evaluation of 3D brain tumor segmentation by DeepIGeoS, GeoS and ITK-SNAP.]{ 
		Quantitative evaluation of 3D brain tumor segmentation by DeepIGeoS, GeoS and ITK-SNAP.} 
	\label{fig:tumor_others_bar}
\end{figure}
Fig.~\ref{fig:tumor_others} shows a visual comparison between GeoS~\cite{Criminisi2008}, ITK-SNAP~\cite{Yushkevich2006} and DeepIGeoS. In the first row, the tumor has a good contrast with the background. All the compared methods achieve very accurate segmentations. In the second row, a lower contrast makes it difficult for the user to identify the tumor boundary. Benefited from the initial tumor boundary that is automatically identified by 3D P-Net, DeepIGeoS outperforms GeoS and ITK-SNAP. Quantitative comparison is presented in Fig.~\ref{fig:tumor_others_bar}. It shows DeepIGeoS achieves higher accuracy compared with GeoS and ITK-SNAP. In addition, the user time for DeepIGeoS is about one third of that for the other two methods. Supplementary video 2 shows more examples of DeepIGeoS for 3D brain tumor segmentation. 

\section{Conclusion}
In this work, we presented a deep learning-based interactive framework for 2D and 3D medical image segmentation. We proposed a P-Net to obtain an initial automatic segmentation and an R-Net to refine the result based on user interactions that are transformed into geodesic distance maps and then integrated into the input of R-Net. We also proposed a resolution-preserving network structure with dilated convolution for dense prediction, and extended the existing RNN-based CRF so that it can learn freeform pairwise potentials and take advantage of user interactions as hard constraints. Segmentation results of the placenta from 2D fetal MRI and brain tumors from 3D FLAIR images show that our proposed method achieves better results than automatic CNNs. It requires far less user time compared with traditional interactive methods and achieves higher accuracy for 3D brain tumor segmentation. The framework can be extended to deal with multiple organs in the future.


%



\ifCLASSOPTIONcompsoc
  \section*{Acknowledgments}
\else
  \section*{Acknowledgment}
\fi

This work was supported through an Innovative Engineering for Health award by the Wellcome Trust (WT101957); Engineering and Physical Sciences Research Council (EPSRC) (NS/A000027/1, EP/H046410/1, EP/J020990/1, EP/K005278), Wellcome/EPSRC [203145Z/16/Z],  the National Institute for Health Research University College London Hospitals Biomedical Research Centre (NIHR BRC UCLH/UCL), the Royal Society [RG160569], a UCL Overseas Research Scholarship and a UCL Graduate Research Scholarship, hardware donated by NVIDIA and of Emerald, a GPU-accelerated High Performance Computer, made available by the Science \& Engineering South Consortium operated in partnership with the STFC Rutherford-Appleton Laboratory.

\ifCLASSOPTIONcaptionsoff
  \newpage
\fi



\bibliographystyle{IEEEtran}
\bibliography{./reference}

\begin{thebibliography}{10}
\providecommand{\url}[1]{#1}
\csname url@samestyle\endcsname
\providecommand{\newblock}{\relax}
\providecommand{\bibinfo}[2]{#2}
\providecommand{\BIBentrySTDinterwordspacing}{\spaceskip=0pt\relax}
\providecommand{\BIBentryALTinterwordstretchfactor}{4}
\providecommand{\BIBentryALTinterwordspacing}{\spaceskip=\fontdimen2\font plus
\BIBentryALTinterwordstretchfactor\fontdimen3\font minus
  \fontdimen4\font\relax}
\providecommand{\BIBforeignlanguage}[2]{{%
\expandafter\ifx\csname l@#1\endcsname\relax
\typeout{** WARNING: IEEEtran.bst: No hyphenation pattern has been}%
\typeout{** loaded for the language `#1'. Using the pattern for}%
\typeout{** the default language instead.}%
\else
\language=\csname l@#1\endcsname
\fi
#2}}
\providecommand{\BIBdecl}{\relax}
\BIBdecl

\bibitem{Sharma2010}
N.~Sharma and L.~M. Aggarwal, ``{Automated medical image segmentation
  techniques.}'' \emph{Journal of medical physics}, vol.~35, no.~1, pp. 3--14,
  2010.

\bibitem{Zhao2013}
F.~Zhao and X.~Xie, ``{An Overview of Interactive Medical Image
  Segmentation},'' \emph{Annals of the BMVA}, vol. 2013, no.~7, pp. 1--22,
  2013.

\bibitem{Boykov2001}
Y.~Boykov and M.-P. Jolly, ``{Interactive Graph Cuts for Optimal Boundary {\&}
  Region Segmentation of Objects in N-D Images},'' in \emph{ICCV}, 2001, pp.
  105--112.

\bibitem{Xu1998}
C.~Xu and J.~L. Prince, ``{Snakes, Shapes, and Gradient Vector Flow},''
  \emph{TIP}, vol.~7, no.~3, pp. 359--369, 1998.

\bibitem{Grady2006a}
L.~Grady, ``{Random walks for image segmentation},'' \emph{PAMI}, vol.~28,
  no.~11, pp. 1768--1783, 2006.

\bibitem{Criminisi2008}
A.~Criminisi, T.~Sharp, and A.~Blake, ``{GeoS: Geodesic Image Segmentation},''
  in \emph{ECCV}, 2008.

\bibitem{Rother2004}
C.~Rother, V.~Kolmogorov, and A.~Blake, ``{"GrabCut": Interactive Foreground
  Extraction Using Iterated Graph Cuts},'' \emph{ACM Trans. on Graphics},
  vol.~23, no.~3, pp. 309--314, 2004.

\bibitem{Wang2016}
G.~Wang, M.~A. Zuluaga, R.~Pratt, M.~Aertsen, T.~Doel, M.~Klusmann, A.~L.
  David, J.~Deprest, T.~Vercauteren, and S.~Ourselin, ``{Slic-Seg: A minimally
  interactive segmentation of the placenta from sparse and motion-corrupted
  fetal MRI in multiple views},'' \emph{Medical Image Analysis}, vol.~34, pp.
  137--147, 2016.

\bibitem{Wang2014a}
B.~Wang, W.~Liu, M.~Prastawa, A.~Irimia, P.~M. Vespa, J.~D.~V. Horn, P.~T.
  Fletcher, and G.~Gerig, ``{4D active cut: An interactive tool for
  pathological anatomy modeling},'' in \emph{ISBI}, 2014.

\bibitem{Girshick2014}
R.~Girshick, J.~Donahue, T.~Darrell, and J.~Malik, ``{Rich Feature Hierarchies
  for Accurate Object Detection and Semantic Segmentation},'' in \emph{CVPR},
  2014.

\bibitem{Long2014}
J.~Long, E.~Shelhamer, and T.~Darrell, ``{Fully Convolutional Networks for
  Semantic Segmentation},'' in \emph{CVPR}, 2015, pp. 3431--3440.

\bibitem{Chen2015iclr}
L.-C. Chen, G.~Papandreou, I.~Kokkinos, K.~Murphy, and A.~L. Yuille,
  ``{Semantic Image Segmentation with Deep Convolutional Nets and Fully
  Connected CRFs},'' in \emph{ICLR}, 2015.

\bibitem{Havaei2016}
M.~Havaei, A.~Davy, D.~Warde-Farley, A.~Biard, A.~Courville, Y.~Bengio, C.~Pal,
  P.-M. Jodoin, and H.~Larochelle, ``{Brain Tumor Segmentation with Deep Neural
  Networks},'' \emph{Medical Image Analysis}, vol.~35, pp. 18--31, 2016.

\bibitem{Kamnitsas2017}
K.~Kamnitsas, C.~Ledig, V.~F.~J. Newcombe, J.~P. Simpson, A.~D. Kane, D.~K.
  Menon, D.~Rueckert, and B.~Glocker, ``{Efficient Multi-Scale 3D CNN with
  Fully Connected CRF for Accurate Brain Lesion Segmentation},'' \emph{Medical
  Image Analysis}, vol.~36, pp. 61--78, 2017.

\bibitem{Onvolutions2016}
F.~Yu and V.~Koltun, ``{Multi-Scale Context Aggregation By Dilated
  Convolutions},'' in \emph{ICLR}, 2016.

\bibitem{Chen2016deeplab}
L.-C. Chen, G.~Papandreou, I.~Kokkinos, K.~Murphy, and A.~L. Yuille,
  ``{DeepLab: Semantic Image Segmentation with Deep Convolutional Nets, Atrous
  Convolution, and Fully Connected CRFs},'' \emph{PAMI}, vol.~PP, no.~99, pp.
  1--1, 2017.

\bibitem{Zheng2015a}
S.~Zheng, S.~Jayasumana, B.~Romera-Paredes, V.~Vineet, Z.~Su, D.~Du, C.~Huang,
  and P.~H.~S. Torr, ``{Conditional Random Fields as Recurrent Neural
  Networks},'' in \emph{ICCV}, 2015.

\bibitem{Abdulkadir2016}
A.~Abdulkadir, S.~S. Lienkamp, T.~Brox, and O.~Ronneberger, ``{3D U-Net :
  Learning Dense Volumetric Segmentation from Sparse Annotation},'' in
  \emph{MICCAI}, 2016.

\bibitem{Lin2016}
D.~Lin, J.~Dai, J.~Jia, K.~He, and J.~Sun, ``{ScribbleSup: Scribble-Supervised
  Convolutional Networks for Semantic Segmentation},'' in \emph{CVPR}, 2016.

\bibitem{Rajchl2016}
M.~Rajchl, M.~Lee, O.~Oktay, K.~Kamnitsas, J.~Passerat-Palmbach, W.~Bai,
  M.~Rutherford, J.~Hajnal, B.~Kainz, and D.~Rueckert, ``{DeepCut: Object
  Segmentation from Bounding Box Annotations using Convolutional Neural
  Networks},'' \emph{TMI}, vol.~PP, no.~99, pp. 1--1, 2016.

\bibitem{Xu2016}
N.~Xu, B.~Price, S.~Cohen, J.~Yang, and T.~Huang, ``{Deep Interactive Object
  Selection},'' in \emph{CVPR}, 2016, pp. 373--381.

\bibitem{Krizhevsky2012}
A.~Krizhevsky, I.~Sutskever, and G.~E. Hinton, ``{ImageNet classification with
  deep convolutional neural networks},'' in \emph{NIPS}, 2012.

\bibitem{Szegedy2015}
C.~Szegedy, W.~Liu, Y.~Jia, P.~Sermanet, S.~Reed, D.~Anguelov, D.~Erhan,
  V.~Vanhoucke, and A.~Rabinovich, ``{Going deeper with convolutions},'' in
  \emph{CVPR}, 2015.

\bibitem{Simonyan2015}
K.~Simonyan and A.~Zisserman, ``{Very deep convolutional networks for
  large-scale image recognition},'' in \emph{ICLR}, 2015.

\bibitem{He2015res}
K.~He, X.~Zhang, S.~Ren, and J.~Sun, ``{Deep Residual Learning for Image
  Recognition},'' in \emph{CVPR}, 2016.

\bibitem{Hefny2015a}
M.~S. Hefny, T.~Okada, M.~Hori, Y.~Sato, and R.~E. Ellis, ``{U-Net:
  Convolutional Networks for Biomedical Image Segmentation},'' in
  \emph{MICCAI}, 2015, pp. 234--241.

\bibitem{Milletari2016}
F.~Milletari, N.~Navab, and S.-A. Ahmadi, ``{V-Net: Fully Convolutional Neural
  Networks for Volumetric Medical Image Segmentation},'' in \emph{IC3DV}, 2016,
  pp. 565--571.

\bibitem{Ondruska2016}
P.~Ondruska, J.~Dequaire, D.~Z. Wang, and I.~Posner, ``{End-to-End Tracking and
  Semantic Segmentation Using Recurrent Neural Networks},'' in \emph{Robotics:
  Science and Systems, Workshop on Limits and Potentials of Deep Learning in
  Robotics}, 2016.

\bibitem{Dai2016}
J.~Dai, K.~He, Y.~Li, S.~Ren, and J.~Sun, ``{Instance-sensitive Fully
  Convolutional Networks},'' in \emph{ECCV}, 2016.

\bibitem{Lea2016}
C.~Lea, R.~Vidal, A.~Reiter, and G.~D. Hager, ``{Temporal Convolutional
  Networks: A Unified Approach to Action Segmentation},'' in \emph{ECCV}, 2016.

\bibitem{Lin2016cvpr_efficient}
G.~Lin, C.~Shen, I.~Reid, and A.~van~dan Hengel, ``{Efficient piecewise
  training of deep structured models for semantic segmentation},'' in
  \emph{CVPR}, 2016.

\bibitem{Pinheiro2014}
P.~Pinheiro and R.~Collobert, ``{Recurrent convolutional neural networks for
  scene labeling},'' in \emph{ICML}, 2014.

\bibitem{Hariharan2015}
B.~Hariharan, P.~Arbel{\'{a}}ez, R.~Girshick, and J.~Malik, ``{Hypercolumns for
  object segmentation and fine-grained localization},'' in \emph{CVPR}, 2015.

\bibitem{Armstrong2007}
C.~J. Armstrong, B.~L. Price, and W.~A. Barrett, ``{Interactive segmentation of
  image volumes with Live Surface},'' \emph{Computers and Graphics (Pergamon)},
  vol.~31, no.~2, pp. 212--229, 2007.

\bibitem{Cates2004}
J.~E. Cates, A.~E. Lefohn, and R.~T. Whitaker, ``{GIST: An interactive,
  GPU-based level set segmentation tool for 3D medical images},'' \emph{Medical
  Image Analysis}, vol.~8, no.~3, pp. 217--231, 2004.

\bibitem{Haider2015}
S.~A. Haider, M.~J. Shafiee, A.~Chung, F.~Khalvati, A.~Oikonomou, A.~Wong, and
  M.~A. Haider, ``{Single-click, semi-automatic lung nodule contouring using
  hierarchical conditional random fields},'' in \emph{ISBI}, 2015.

\bibitem{Bai2007}
X.~Bai and G.~Sapiro, ``{A Geodesic Framework for Fast Interactive Image and
  Video Segmentation and Matting},'' in \emph{ICCV}, oct 2007.

\bibitem{Barinova2012}
O.~Barinova, R.~Shapovalov, S.~Sudakov, and A.~Velizhev, ``{Online Random
  Forest for Interactive Image Segmentation},'' in \emph{EEML}, 2012.

\bibitem{Luengo2017}
I.~Luengo, M.~C. Darrow, M.~C. Spink, Y.~Sun, W.~Dai, C.~Y. He, W.~Chiu,
  T.~Pridmore, A.~W. Ashton, E.~M. Duke, M.~Basham, and A.~P. French,
  ``{SuRVoS: Super-Region Volume Segmentation workbench},'' \emph{Journal of
  Structural Biology}, vol. 198, no.~1, pp. 43--53, 2017.

\bibitem{Kohli2013}
P.~Kohli, J.~Shotton, and A.~Criminisi, ``{GeoF Geodesic Forests for Learning
  Coupled Predictors},'' in \emph{CVPR}, 2013.

\bibitem{Boykov2004}
Y.~Boykov and V.~Kolmogorov, ``{An experimental comparison of min-cut/max-flow
  algorithms for energy minimization in vision},'' \emph{PAMI}, vol.~26, no.~9,
  pp. 1124--1137, 2004.

\bibitem{Yuan2010}
J.~Yuan, E.~Bae, and X.~C. Tai, ``{A study on continuous max-flow and min-cut
  approaches},'' in \emph{CVPR}, 2010.

\bibitem{Payet2010}
N.~Payet and S.~Todorovic, ``{(RF){\^{}}2 -- Random Forest Random Field},''
  \emph{NIPS}, vol.~1, pp. 1885--1893, 2010.

\bibitem{Krahenbuhl2011}
P.~Kr{\"{a}}henb{\"{u}}hl and V.~Koltun, ``{Efficient inference in fully
  connected CRFs with gaussian edge potentials},'' in \emph{NIPS}, 2011.

\bibitem{Szummer2008}
M.~Szummer, P.~Kohli, and D.~Hoiem, ``{Learning CRFs using graph cuts},'' in
  \emph{ECCV}, 2008.

\bibitem{Blaschko2014}
J.~I. Orlando and M.~Blaschko, ``{Learning Fully-Connected CRFs for Blood
  Vessel Segmentation in Retinal Images},'' in \emph{MICCAI}, 2014.

\bibitem{Domke2013}
J.~Domke, ``{Learning graphical model parameters with approximate marginal
  inference},'' \emph{PAMI}, vol.~35, no.~10, pp. 2454--2467, 2013.

\bibitem{Krahenbuhl2013a}
P.~Kr{\"{a}}henb{\"{u}}hl and V.~Koltun, ``{Parameter Learning and Convergent
  Inference for Dense Random Fields},'' in \emph{ICML}, 2013, pp. 1--9.

\bibitem{Adams2010}
A.~Adams, J.~Baek, and M.~A. Davis, ``{Fast high-dimensional filtering using
  the permutohedral lattice},'' \emph{Computer Graphics Forum}, vol.~29, no.~2,
  pp. 753--762, 2010.

\bibitem{Vemulapalli2016}
R.~Vemulapalli, O.~Tuzel, M.-y. Liu, and R.~Chellappa, ``{Gaussian Conditional
  Random Field Network for Semantic Segmentation},'' in \emph{CVPR}, 2016.

\bibitem{Liu2014}
F.~Liu, C.~Shen, and G.~Lin, ``{Deep Convolutional Neural Fields for Depth
  Estimation from a Single Image},'' in \emph{CVPR}, 2014.

\bibitem{Kirillov2016}
A.~Kirillov, S.~Zheng, D.~Schlesinger, W.~Forkel, A.~Zelenin, P.~Torr, and
  C.~Rother, ``{Efficient Likelihood Learning of a Generic CNN-CRF Model for
  Semantic Segmentation},'' in \emph{ACCV}, 2016.

\bibitem{Wang2016a}
G.~Wang, M.~A. Zuluaga, R.~Pratt, M.~Aertsen, T.~Doel, M.~Klusmann, A.~L.
  David, J.~Deprest, T.~Vercauteren, and S.~Ourselin, ``{Dynamically Balanced
  Online Random Forests for Interactive Scribble-Based Segmentation},'' in
  \emph{MICCAI}, 2016.

\bibitem{Kolmogorov2004}
V.~Kolmogorov and R.~Zabih, ``{What Energy Functions Can Be Minimized via Graph
  Cuts?}'' \emph{PAMI}, vol.~26, no.~2, pp. 147--159, 2004.

\bibitem{Han2011ipmi}
D.~Han, J.~Bayouth, Q.~Song, A.~Taurani, M.~Sonka, J.~Buatti, and X.~Wu,
  ``{Globally optimal tumor segmentation in PET-CT images: A graph-based
  co-segmentation method},'' in \emph{IPMI}, 2011.

\bibitem{Jia2014}
Y.~Jia, E.~Shelhamer, J.~Donahue, S.~Karayev, J.~Long, R.~Girshick,
  S.~Guadarrama, and T.~Darrell, ``{Caffe: Convolutional Architecture for Fast
  Feature Embedding},'' in \emph{ACMICM}, 2014.

\bibitem{Abadi2016}
M.~Abadi, P.~Barham, J.~Chen, Z.~Chen, A.~Davis, J.~Dean, M.~Devin,
  S.~Ghemawat, G.~Irving, M.~Isard, M.~Kudlur, J.~Levenberg, R.~Monga,
  S.~Moore, D.~G. Murray, B.~Steiner, P.~Tucker, V.~Vasudevan, P.~Warden,
  M.~Wicke, Y.~Yu, X.~Zheng, and G.~Brain, ``{TensorFlow: A System for
  Large-Scale Machine Learning TensorFlow: A system for large-scale machine
  learning},'' in \emph{OSDI}, 2016, pp. 265--284.

\bibitem{Li2017}
W.~Li, G.~Wang, L.~Fidon, S.~Ourselin, M.~J. Cardoso, and T.~Vercauteren, ``{On
  the Compactness , Efficiency , and Representation of 3D Convolutional
  Networks : Brain Parcellation as a Pretext Task},'' in \emph{IPMI}, 2017.

\bibitem{Yushkevich2006}
P.~A. Yushkevich, J.~Piven, H.~C. Hazlett, R.~G. Smith, S.~Ho, J.~C. Gee, and
  G.~Gerig, ``{User-guided 3D active contour segmentation of anatomical
  structures: Significantly improved efficiency and reliability},''
  \emph{NeuroImage}, vol.~31, no.~3, pp. 1116--1128, 2006.

\bibitem{Deprest2010}
J.~A. Deprest, A.~W. Flake, E.~Gratacos, Y.~Ville, K.~Hecher, K.~Nicolaides,
  M.~P. Johnson, F.~I. Luks, N.~S. Adzick, and M.~R. Harrison, ``{The Making of
  Fetal Surgery},'' \emph{Prenatal Diagnosis}, vol.~30, no.~7, pp. 653--667,
  2010.

\bibitem{Alansary2016}
A.~Alansary, K.~Kamnitsas, A.~Davidson, M.~Rajchl, C.~Malamateniou,
  M.~Rutherford, J.~V. Hajnal, B.~Glocker, D.~Rueckert, and B.~Kainz, ``{Fast
  Fully Automatic Segmentation of the Human Placenta from Motion Corrupted
  MRI},'' in \emph{MICCAI}, 2016.

\bibitem{Keraudren2014}
K.~Keraudren, M.~Kuklisova-Murgasova, V.~Kyriakopoulou, C.~Malamateniou, M.~A.
  Rutherford, B.~Kainz, J.~V. Hajnal, and D.~Rueckert, ``{Automated Fetal Brain
  Segmentation from 2D MRI Slices for Motion Correction},'' \emph{NeuroImage},
  vol. 101, no. 1 November 2014, pp. 633--643, 2014.

\bibitem{Menze2015a}
B.~H. Menze, A.~Jakab, S.~Bauer, J.~Kalpathy-Cramer, K.~Farahani, J.~Kirby,
  Y.~Burren, N.~Porz, J.~Slotboom, R.~Wiest, L.~Lanczi, E.~Gerstner, M.~A.
  Weber, T.~Arbel, B.~B. Avants, N.~Ayache, P.~Buendia, D.~L. Collins,
  N.~Cordier, J.~J. Corso, A.~Criminisi, T.~Das, H.~Delingette,
  {\c{C}}.~Demiralp, C.~R. Durst, M.~Dojat, S.~Doyle, J.~Festa, F.~Forbes,
  E.~Geremia, B.~Glocker, P.~Golland, X.~Guo, A.~Hamamci, K.~M. Iftekharuddin,
  R.~Jena, N.~M. John, E.~Konukoglu, D.~Lashkari, J.~A. Mariz, R.~Meier,
  S.~Pereira, D.~Precup, S.~J. Price, T.~R. Raviv, S.~M. Reza, M.~Ryan,
  D.~Sarikaya, L.~Schwartz, H.~C. Shin, J.~Shotton, C.~A. Silva, N.~Sousa,
  N.~K. Subbanna, G.~Szekely, T.~J. Taylor, O.~M. Thomas, N.~J. Tustison,
  G.~Unal, F.~Vasseur, M.~Wintermark, D.~H. Ye, L.~Zhao, B.~Zhao, D.~Zikic,
  M.~Prastawa, M.~Reyes, and K.~{Van Leemput}, ``{The Multimodal Brain Tumor
  Image Segmentation Benchmark (BRATS)},'' \emph{TMI}, vol.~34, no.~10, pp.
  1993--2024, 2015.

\bibitem{Fidon2017a}
L.~Fidon, W.~Li, L.~C. Garcia-peraza herrera, J.~Ekanayake, N.~Kitchen,
  S.~Ourselin, and T.~Vercauteren, ``{Scalable multimodel convolutional
  networks for brain tumour segmentation},'' in \emph{MICCAI}, 2017.

\end{thebibliography}
%



%

\begin{IEEEbiography}[{\includegraphics[width=1in,height=1.25in,clip,keepaspectratio]{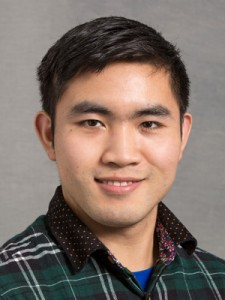}}]
	{Guotai Wang}
obtained his Bachelor and Master degree of Biomedical Engineering in Shanghai Jiao Tong University in 2011 and 2014 respectively. 
He then joined Translational Imaging Group in UCL as a PhD student, working on image segmentation for fetal surgical planning. He won the UCL Overseas Research Scholarship and UCL Graduate Research Scholarship. His research interests include image segmentation, computer vision and deep learning.

\end{IEEEbiography}
\begin{IEEEbiography}[{\includegraphics[width=1in,height=1.25in,clip,keepaspectratio]{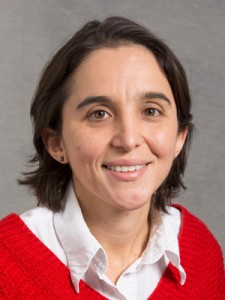}}]
	{Maria A. Zuluaga}
obtained her PhD degree from Université Claude Bernard Lyon. 
After a year as a postdoctoral fellow at the European Synchrotron Radiation Facility (Grenoble, France), she joined CMIC, in March 2012, as a Research Associate to work on cardiovascular image analysis and computer-aided diagnosis (CAD) of cardiovascular pathologies. Since August 2014, she has been a part of the Guided Instrumentation for Fetal Therapy and Surgery (GIFT-Surg) project as a senior research associate.
\end{IEEEbiography}
\begin{IEEEbiography}[{\includegraphics[width=1in,height=1.25in,clip,keepaspectratio]{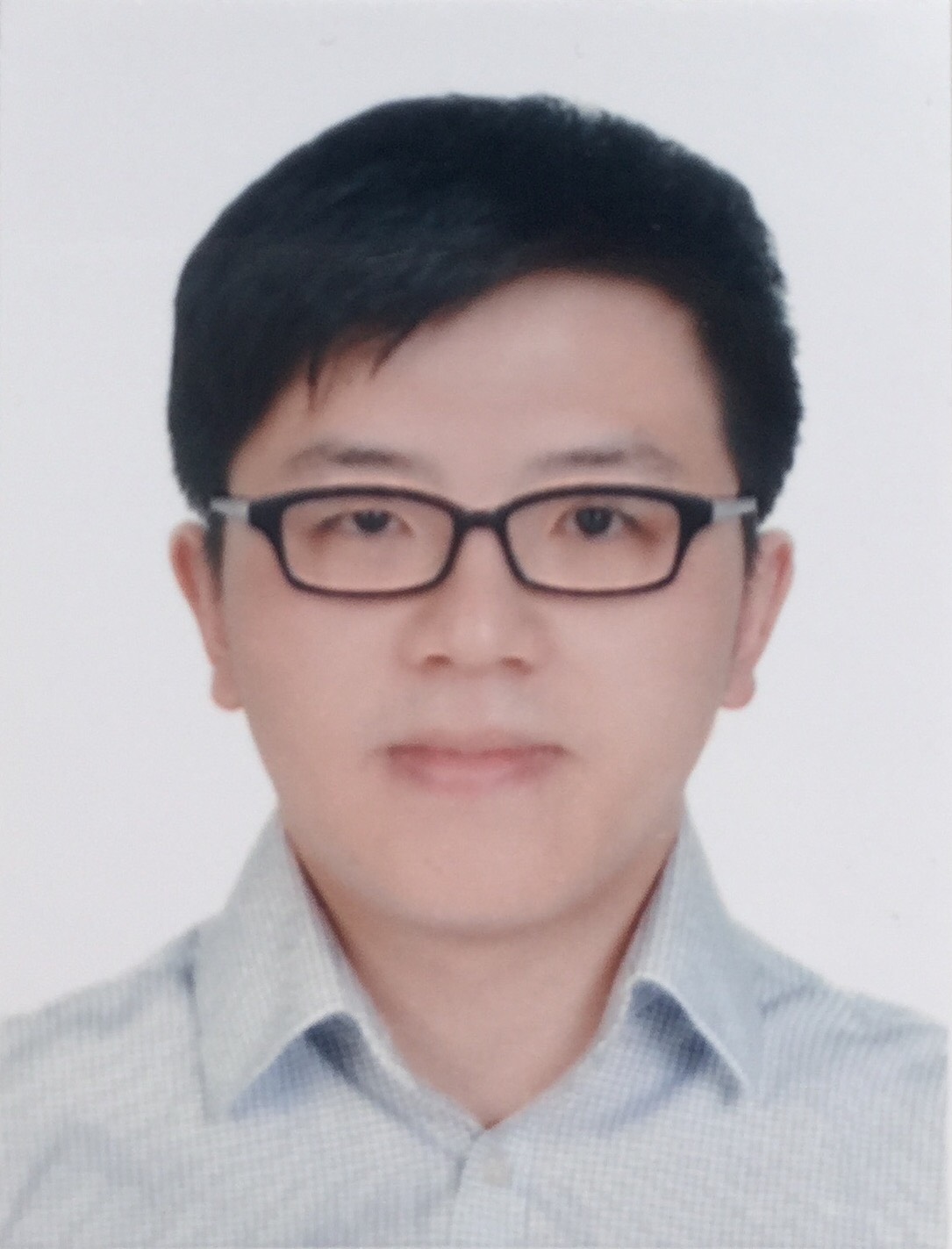}}]
	{Wenqi Li} is a Research Associate in the Guided Instrumentation for Fetal Therapy and Surgery (GIFT-Surg) project. His main research interests are in anatomy detection and segmentation for presurgical evaluation and surgical planning. Before joining TIG, he obtained a BSc degree in Computer Science from the University of Science and Technology Beijing in 2010, and then an MSc degree in Computing with Vision and Imaging from the University of Dundee in 2011. In 2015, he completed his PhD in the Computer Vision and Image Processing group at the University of Dundee.
\end{IEEEbiography}

\begin{IEEEbiography}[{\includegraphics[width=1in,height=1.25in,clip,keepaspectratio]{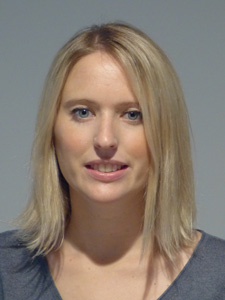}}]
	{Rosalind Pratt} is a Clinical Academic Training Fellow at UCL Institute for Women's Health. She studied medicine at the University of Leeds, before starting clinical training in London, where she is undertaking specialist training in Obstetrics and Gynaecology. Her main research focus is in novel imaging of the human placenta.
\end{IEEEbiography}

\begin{IEEEbiography}[{\includegraphics[width=1in,height=1.25in,clip,keepaspectratio]{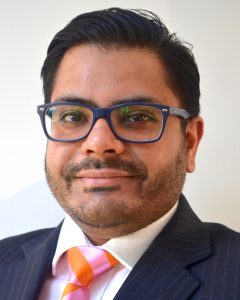}}]
	{Premal A. Patel} is a Consultant Paediatric Interventional Radiologist at Great Ormond Street Hospital for Children NHS Foundation Trust and undertakes a variety of image guided procedures in children.  He is also a Clinical Training Fellow within the Translational Imaging Group at UCL. He studied medicine at St Bartholomew’s and the Royal London Hospital School of Medicine and Dentistry, University of London. Following a period of training in Surgery and Paediatric Surgery in London, he undertook training in Clinical Radiology in Southampton and subsequently undertook Fellowships in Paediatric Interventional Radiology at Great Ormond Street Hospital for Children, London, UK and The Hospital for Sick Children (SickKids), University of Toronto, Canada. 
\end{IEEEbiography}

\begin{IEEEbiography}[{\includegraphics[width=1in,height=1.25in,clip,keepaspectratio]{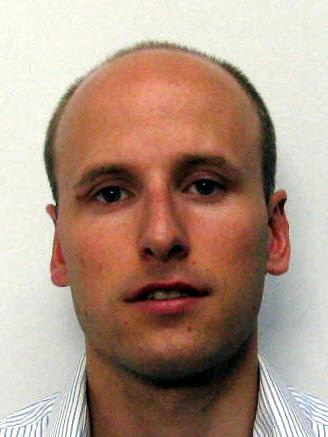}}]
	{Michael Aertsen} is a Consultant Pediatric Radiologist at University Hospitals of Leuven. He studied medecine at the University of Hasselt and the Katholieke Universiteit Leuven. He is specialized in fetal MRI and his main research focus is the fetal brain development with advanced MRI techniques.
\end{IEEEbiography}

\begin{IEEEbiography}[{\includegraphics[width=1in,height=1.25in,clip,keepaspectratio]{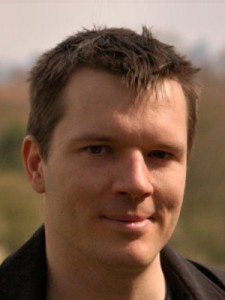}}]
	{Tom Doel}is a Research Associate in the Translational Imaging Group at UCL and an Honorary Research Fellow at UCL Hospitals NHS Foundation Trust. His PhD at the University of Oxford developed computational techniques for multi-modal medical image segmentation, registration, analysis and modelling. He holds an MSc in applied mathematics from the University of Bath and an MPhys in mathematical physics from the University of Edinburgh. He is a professional software engineer and worked for Barco Medical Imaging Systems to develop clinical radiology software. His current research is on novel algorithm development and the robust design and architecture of clinically-focussed software for surgical planning and image-guided surgery as part of the GIFT-Surg project.
\end{IEEEbiography}


\begin{IEEEbiography}[{\includegraphics[width=1in,height=1.25in,clip,keepaspectratio]{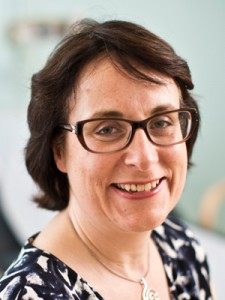}}]
	{Anna L. David} is a Professor and Honorary Consultant in Obstetrics and Maternal Fetal Medicine at Institute for Women's Health, University College London (UCL), London. She has a clinical practice at UCLH in fetal medicine, fetal therapy and obstetrics. Her main research is in translational medicine. She is Head of the Research Department of Maternal Fetal Medicine at UCL Institute for Women's Health. 
\end{IEEEbiography}

\begin{IEEEbiography}[{\includegraphics[width=1in,height=1.25in,clip,keepaspectratio]{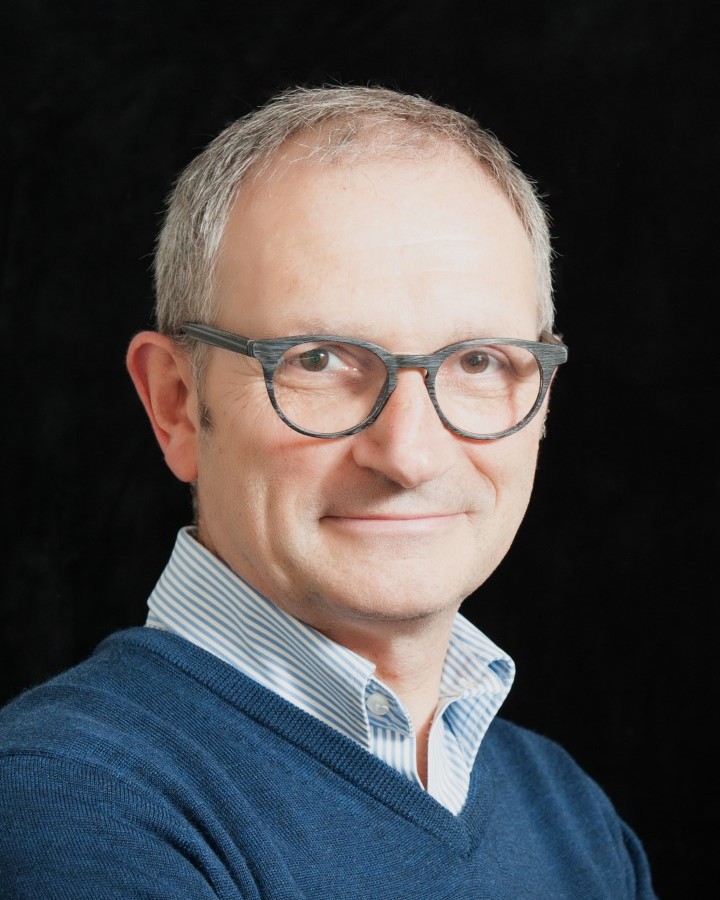}}]
	{Jan Deprest} is a Professor of Obstetrics and Gynaecology at the Katholieke Universiteit Leuven and Consultant Obstetrician Gynaecologist at the University Hospitals Leuven (Belgium). He is currently the academic chair of his department and the director of the Centre for Surgical Technologies at the Faculty of Medicine. He established the Eurofoetus consortium, which is dedicated to the development of instruments and techniques for minimally invasive fetal and placental surgery. 
\end{IEEEbiography}


\begin{IEEEbiography}[{\includegraphics[width=1in,height=1.25in,clip,keepaspectratio]{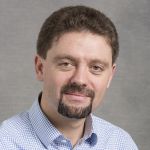}}]
	{S\'ebastien Ourselin}
	is 
	Director of the EPSRC Centre for Doctoral Training (CDT) in Medical Imaging, Head of the Translational Imaging Group (TIG) as part of the Centre for Medical Image Computing (CMIC), and Professor of Medical Image Computing at UCL. His core skills are in medical image analysis, software engineering, and translational medicine. He is best known for his work on image registration and segmentation, its exploitation for robust image-based biomarkers in neurological conditions, as well as for his development of image-guided surgery systems. 
\end{IEEEbiography}

\begin{IEEEbiography}[{\includegraphics[width=1in,height=1.25in,clip,keepaspectratio]{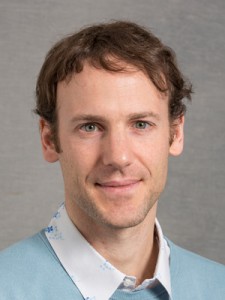}}]
	{Tom Vercauteren}
is a Senior Lecturer within the Translational Imaging Group at UCL. 
He is a graduate from Columbia University and Ecole Polytechnique and obtained his PhD from Inria Sophia Antipolis. His main research focus is on the development of innovative interventional imaging systems and their translation to the clinic. One key driving force of his work is the exploitation of image computing and the knowledge of the physics of acquisition to move beyond the initial limitations of the medical imaging devices that are developed or used in the course of his research. 
\end{IEEEbiography}




\end{document}